\documentclass[11pt]{article}
\usepackage[utf8]{inputenc}
\usepackage[T1]{fontenc}
\usepackage[margin={1.5in}]{geometry}

\usepackage{mathtools,amssymb,pifont,adjustbox,graphicx,color,dsfont,multirow,multicol,tabularx,pifont,booktabs,lipsum}
\usepackage{bigdelim,bbm,url,etoolbox,xr,amsmath,amsfonts,amsthm,mathabx,indentfirst,comment,titlesec,secdot,enumerate,psfrag,epsf}

\usepackage[para,online,flushleft]{threeparttable}
\usepackage[most]{tcolorbox}
\usepackage[labelsep=period]{caption}
\usepackage{subcaption}
\usepackage[hidelinks,bookmarks=false]{hyperref}
\usepackage[page,title,titletoc,header]{appendix}
\usepackage[bottom]{footmisc}

\usepackage{setspace}
\numberwithin{equation}{section}
\numberwithin{table}{section}
\numberwithin{figure}{section}
\allowdisplaybreaks

\usepackage[style=apa,sortcites=false,uniquename=init,backend=biber,natbib=true,hyperref=true,maxbibnames=99,mincitenames=1, maxcitenames=2, citetracker]{biblatex}

\renewbibmacro{in:}{}
\newcommand{\one}{\mathbf{1}}
\newcolumntype{L}[1]{>{\raggedright\arraybackslash}p{#1}}
\newcolumntype{C}[1]{>{\centering\arraybackslash}p{#1}}
\newcolumntype{R}[1]{>{\raggedleft\arraybackslash}p{#1}}

\author{Zhenhua Wang, Olanrewaju Akande, Jason Poulos and Fan Li \footnote{Zhenhua Wang is PhD student in the Department of Statistics, University of Missouri, Columbia, MO 65211 (E-mail: \href{mailto:zhenhua.wang@mail.missouri.edu}{zhenhua.wang@mail.missouri.edu}); Olanrewaju Akande is research scientist at Meta Platforms, Inc. (E-mail: \href{mailto:akandelanre13@gmail.com}{akandelanre13@gmail.com}); Jason Poulos is Postdoctoral Associate in the Department of Health Care Policy, Harvard Medical School, Boston, MA (E-mail: \href{mailto:poulos@hcp.med.harvard.edu}{poulos@hcp.med.harvard.edu}); and Fan Li is Professor in the Department of Statistical Science, Box 90251, Duke University, Durham, NC 27708 (E-mail: \href{mailto:fl35@duke.edu}{fl35@duke.edu}).}}
\title{Are deep learning models superior for missing data imputation in surveys? Evidence from an empirical comparison}
\date{}

\begin{document}
\maketitle
\thispagestyle{empty}

\begin{abstract}
Multiple imputation (MI) is a popular approach for dealing with missing data arising from non-response in sample surveys. Multiple imputation by chained equations (MICE) is one of the most widely used MI algorithms for multivariate data, but it lacks theoretical foundation and is computationally intensive. Recently, missing data imputation methods based on deep learning models have been developed with encouraging results in small studies. However, there has been limited research on evaluating their performance in realistic settings compared to MICE, particularly in big surveys. We conduct extensive simulation studies based on a subsample of the American Community Survey to compare the repeated sampling properties of four machine learning based MI methods: MICE with classification trees, MICE with random forests, generative adversarial imputation networks, and multiple imputation using denoising autoencoders. We find the deep learning imputation methods are superior to MICE in terms of computational time. However, with the default choice of hyperparameters in the common software packages, MICE with classification trees consistently outperforms, often by a large margin, the deep learning imputation methods in terms of bias, mean squared error, and coverage under a range of realistic settings.
\end{abstract}

\setcounter{page}{1}
\section{Introduction} \label{sec:intro}

Many sample surveys suffer from missing data, arising from unit nonresponse, where a subset of participants do not complete the survey, or item nonresponse, where missing values are concentrated on particular questions. In opinion polls, nonresponse may reflect either refusal to reveal a preference or lack of a preference \citep{de2003prevention}. If not properly handled, missing data patterns can lead to biased statistical analyses, especially when there are systematic differences between the observed data and the missing data \citep{Rubin1976,LittleRubin2019}. Complete case analysis on units with completely observed data is often infeasible and may lead to large bias in most situations \citep{LittleRubin2019}. As a result, many analysts account for the missing data by imputing missing values and then proceeding as if the imputed values are true values.

Multiple imputation (MI) \citep{Rubin1987} is a popular approach for handling missing values. In MI, an analyst creates $L > 1$ completed datasets by replacing the missing values in the sample data with plausible draws generated from the predictive distribution of probabilistic models based on the observed data. In each completed dataset, the analyst can then compute sample estimates for population estimands of interest, and combine the sample estimates across all $L$ datasets using MI inference methods developed by \citet{Rubin1987}, and more recently, \citet{Rubin1996,BarnardMeng1999, ReiterRaghunathan2007}, and \citet{HarelZhou2007}. In MI, the estimated variance of an estimand consists of both within-imputation and between-imputation variances, and thus takes into account the inherent variability of the imputed values. Note that in survey studies, single imputation, e.g. via matching or regression, remains to be common for dealing with missing data, where the variance is estimated via the delta method or resampling methods \citep{chen2019recent,haziza2020variance}. 

\subsection{Model-based imputation}
There are two general modeling strategies for MI. The first strategy, known as \emph{joint modeling} (JM), is to specify a joint distribution for all variables in the data, and then generate imputations from the implied conditional (predictive) distributions of the variables with missing values \citep{Schafer1997a}. The JM strategy aligns with the theoretical foundation of \citet{Rubin1987}, but it can be challenging to specify a joint model with high-dimensional variables of different types. Indeed, most popular JM approaches, such as ``PROC MI'' in SAS \citep{yuan2011multiple}, and ``AMELIA'' \citep{honaker2011amelia} and ``norm'' in \textsf{R} \citep{Schafer1997a}, make a simplifying assumption that the data follow multivariate Gaussian distributions, even for categorical variables, which can lead to bias \citep{horton2003potential}. Recent research developed flexible JM models based on advanced Bayesian nonparametric models such as Dirichlet Process mixtures \citep{manrique2014bayesian,murray2016multiple}. However, these methods are computationally expensive, and often struggle to scale up to high-dimensional cases.

The second strategy is called \emph{fully conditional specification} (FCS, \citet{VanBuuren2006}), where one separately specifies a univariate conditional distribution for each variable with missing values given all the other variables and imputes the missing values variable-by-variable iteratively, akin to a Gibbs sampler. The most popular FCS method is multiple imputation by chained equations (MICE) \citep{mice2011}, usually implemented with specifying generalized linear models (GLMs) for the univariate conditional distributions \citep{raghunathan2001multivariate, royston2011multiple, su2011multiple}. 
Recent research indicates that specifying the conditional models by classification and regression trees (CART, \citet{breiman:1984, burgreit10}) comprehensively outperforms MICE with GLM \citep{akande2017empirical}. A natural extension of MICE with CART is to use ensemble tree methods such as random forests, rather than a single tree \citep{breiman:2001,DooverEtAl2014}.

MICE is appealing in large-scale survey data because it is simple and flexible in imputing different types of variables. However, MICE has a key theoretical drawback that the specified conditional distributions may be incompatible, that is, they do not correspond to a joint distribution \citep{ArnoldPress1989,GelmanSpeed1993, Li2012}. Despite this drawback, MICE works remarkably well in real applications and numerous simulations have demonstrated it outperforms many theoretically sound JM-based methods; see \cite{van2018flexible} for case studies. However, MICE is also computationally intensive \citep{white2011} and generally cannot be parallelized. Moreover, popular software packages for implementing MICE with GLMs, e.g. \texttt{mice} in \textsf{R} \citep{mice2011}, often crash in settings with high dimensional non-continuous variables, e.g., categorical variables with many categories \citep{akande2017empirical}.  

\subsection{Imputation with deep learning models}
Recent advances in deep learning greatly expand the scope of complex models for high-dimensional data. This advancement brings the hope that a new generation of missing data imputation methods based on deep learning models may address the theoretical and computational limitations of existing statistical methods. For example, deep generative models such as generative adversarial networks (GANs, \citet{goodfellow2014gan}) are naturally suitable for producing multiple imputations because they are designed to generate data that resemble the observed data as much as possible. A method in this stream is the generative adversarial imputation network (GAIN) of \citet{yoon2018gain}. Multiple imputation using denoising autoencoders (MIDA, \citet{gondara2017mida,lu2020multiple}), is another generative method based on deep neural networks trained on corrupted input data in order to force the networks to learn a useful low-dimensional representation of the input data, rather than its identity function \citep{Vincent2008, vincent2010stacked}. Several methods have been proposed for missing value imputation in time-series data using variational autoencoders \citep{fortuin2019gpvae} or recurrent neural networks \citep{lipton2016modeling,monti2017geometric,cao2018brits,che2018recurrent,yoon2018estimating}. 

Deep learning based MI methods have several advantages, at least theoretically, over the traditional statistical models, including (i) they avoid making distributional assumptions; (ii) can readily handle mixed data types; (iii) can model nonlinear relationships between variables; (iv) are expected to perform well in high-dimensional settings; and (v) can leverage graphics processing unit (GPU) power for faster computation.
Several papers report encouraging performance of deep learning based MI methods compared to MICE \citep[e.g.][]{yoon2018gain}. However, such conclusions are made based on limited evidence. First, the studies are usually based on small simulations or several well-studied public ``benchmark'' datasets, such as those described in Section \ref{sec:benchmark}, which do not resemble survey data. Second, the evaluations are usually based on a few overall performance metric, e.g., the overall predictive mean squared error or accuracy. Such metrics may not give a full picture of the comparisons and sometimes can be even misleading, as will be illustrated later. Third, given the uncertainty of the missing data process, it is crucial to examine the repeated sampling properties of imputation methods, but these have been rarely evaluated. Finally, hyperparameter tuning is crucial for machine learning models and different tuning can result in dramatically different results, but few details are provided on hyperparameter tuning and its consequences on the performance of imputation methods.

Motivated by these limitations, in this paper we carry out extensive simulations based on real survey data to evaluate MI methods with a range performance metrics. Specifically, we conduct simulations based on a subsample from the American Community Survey to compare repeated sampling properties of four aforementioned MI methods: MICE with CART (MICE-CART), MICE with random forests (MICE-RF), GAIN, and MIDA. We find that deep learning based MI methods are superior to MICE in terms of computational time. However, MICE-CART consistently outperforms, often by a large margin, the deep learning methods in terms of bias, mean squared error, and coverage, under a range of realistic settings. This contradicts previous findings in the machine learning literature, and raises questions on the appropriate metrics for evaluating imputation methods. It also highlights the importance of assessing repeated-sampling properties of imputation methods. Though we focus on multiple imputation in this paper, we note that the aforementioned MI methods are readily applicable to generate single imputation when $L$ is set to 1. Extensive empirical evidences suggest that the within-imputation variance usually dominates the between-imputation variance in MI. As such, we expect the patterns between different imputation methods observed here also stand if these methods are used for single imputation.

The remainder of this article is organized as follows. In Section \ref{sec:methods}, we review the four MI methods used in our evaluation. In Section \ref{sec:framework}, we describe a framework with several metrics for evaluating imputation methods. In Section \ref{sec:evaluation}, we describe the simulation design and results with large-scale survey data, and in Section \ref{sec:benchmark} we summarize evaluation results on the benchmark datasets used in machine learning literature. Finally, in Section \ref{sec:discussion}, we conclude with a practical guide for implementation in real applications.

\section{Missing Data Imputation Methods} \label{sec:methods}
We first introduce notation. Consider a sample with $n$ units, each of which is associated with $p$ variables. Let $Y_{ij}$ be the value of variable $j$ for individual $i$, where $j=1, \dots, p$ and $i=1, \dots, n$. Here, $Y$ can be continuous, binary, categorical or mixed binary-continuous. For each individual $i$, let ${\bf Y}_i = (Y_{i1}, \dots, Y_{ip})$. For each variable $j$, let ${\bf Y}_j = (Y_{1j}, \dots, Y_{nj})$. Let ${\bf Y} = ({\bf Y}_1, \ldots, {\bf Y}_n)$ be the $n \times p$ matrix comprising the data for all records included in the sample.  We write ${\bf Y} = ({\bf Y}_{\textrm{obs}}, {\bf Y}_{\textrm{mis}})$, where ${\bf Y}_{\textrm{obs}}$ and ${\bf Y}_{\textrm{mis}}$ are respectively the observed and missing parts of ${\bf Y}$.  We write ${\bf Y}_{\textrm{mis}} = ({\bf Y}_{\textrm{mis},1}, \dots, {\bf Y}_{\textrm{mis},p})$, where ${\bf Y}_{\textrm{mis},j}$ represents all missing values for variable $j$, with $j=1, \dots, p$.  Similarly, we write ${\bf Y}_{\textrm{obs}} = ({\bf Y}_{\textrm{obs},1}, \dots, {\bf Y}_{\textrm{obs},p})$ for the corresponding observed data.

In MI, the analyst generates values of the missing data ${\bf Y}_{\textrm{mis}}$ using pre-specified models estimated with ${\bf Y}_{\textrm{obs}}$, resulting in a completed dataset. The analyst then repeats the process to generate $L$ completed datasets, $\{{\bf Y}^{(l)}: l= 1, \dots, L\}$, that are available for inference or dissemination. For inference, the analyst can compute sample estimates for population estimands in each completed dataset $\textbf{Y}^{(l)}$, and combine them using MI inference rules developed by \citet{Rubin1987}, which will be reviewed in Section \ref{sec:framework}.

\subsection{MICE with classification tree models}
Under MICE, the analyst begins by specifying a separate univariate conditional model for each variable with missing values. The analyst then specifies an order to iterate through the sequence of the conditional models, when doing imputation. We write the ordered list of the variables as $(\textbf{Y}_{(1)}, \dots, \textbf{Y}_{(p)})$. Next, the analyst initializes each ${\bf Y}_{\textrm{mis},(j)}$. The most popular options are to sample from (i) the marginal distribution of the corresponding ${\bf Y}_{\textrm{obs},(j)}$, or (ii) the conditional distribution of ${\bf Y}_{(j)}$ given all the other variables, constructed using only available cases.

After initialization, the MICE algorithm follows an iterative process that cycles through the sequence of univariate models. For each variable $j$ at each iteration $t$, one fits the conditional model $(\textbf{Y}_{(j)} | \textbf{Y}_{\textrm{obs},(j)}, \{\textbf{Y}_{(k)}^{(t)}: k<j\}, \{\textbf{Y}_{(k)}^{(t-1)}: k>j\})$. Next, one replaces ${\bf Y}_{\textrm{mis},(j)}^{(t)}$ with draws from the implied model $({\bf Y}_{\textrm{mis},(j)}^{(t)} | \textbf{Y}_{\textrm{obs},(j)}, \{\textbf{Y}_{(k)}^{(t)}: k<j\}, \{\textbf{Y}_{(k)}^{(t-1)}: k>j\})$. The iterative process continues for $T$ total iterations until convergence, and the values at the final iteration make up a completed dataset ${\bf Y}^{(l)} = ({\bf Y}_{\textrm{obs}}, {\bf Y}_{\textrm{mis}}^{(T)})$. The entire process is then repeated $L$ times to create the $L$ completed datasets. We provide pseudocode detailing each step of the MICE algorithm in the supplementary material. 

Under MICE-CART, the analyst uses CART \citep{breiman:1984} for the univariate conditional models in the MICE algorithm. CART follows a decision tree structure that uses recursive binary splits to partition the predictor space into distinct non-overlapping regions. The top of the tree often represents its root and each successive binary split divides the predictor space into two new branches as one moves down the tree. The splitting criterion at each leaf is usually chosen to minimize an information theoretic entropy measure. Splits that do not decrease the lack of fit by an reasonable amount based on a set threshold are pruned off. The tree is then built until a stopping criterion is met; e.g., minimum number of observations in each leaf.

Once the tree has been fully constructed, one generates ${\bf Y}_{\textrm{mis},(j)}^{(t)}$ by traversing down the tree to the appropriate leaf using the combinations in $(\{\textbf{Y}_k^{(t)}: k<j\}, \{\textbf{Y}_k^{(t-1)}: k>j\})$, and then sampling from the $\textbf{Y}_{(j)}^{\textrm{obs}}$ values in that leaf. That is, given any combination in $(\{\textbf{Y}_k^{(t)}: k<j\}, \{\textbf{Y}_k^{(t-1)}: k>j\})$, one uses the proportion of values of $\textbf{Y}_{j}^{\textrm{obs}}$ in the corresponding leaf to approximate the conditional distribution $(\textbf{Y}_{(j)} | \textbf{Y}_{\textrm{obs},(j)}, \{\textbf{Y}_{(k)}^{(t)}: k<j\}, \{\textbf{Y}_{(k)}^{(t-1)}: k>j\})$. The iterative process again continues for $T$ total iterations, and the values at the final iteration make up a completed dataset.

MICE-RF instead uses random forests for the univariate conditional models in MICE \citep[e.g.,][]{missforest, Shah:2014}. Random forests \citep{randomForest,breiman:2001} is an ensemble tree method which builds multiple decision trees to the data, instead of a single tree like CART. Specifically, random forests constructs multiple decision trees using bootstrapped samples of the original, and only uses a sample of the predictors for the recursive partitions in each tree. This approach can reduce the prevalence of unstable trees as well as the correlation among individual trees significantly, since it prevents the same variables from dominating the partitioning process across all trees. Theoretically, this decorrelation should result in predictions with less variance \citep{HastieESL:2009}. 

For imputation, the analyst first trains a random forests model for each $\textbf{Y}_{(j)}$ using available cases, given all other variables. Next, the analyst generates predictions for ${\bf Y}_{\textrm{mis},j}$ under that model. Specifically, for any categorical $\textbf{Y}_{(j)}$, and given any particular combination in $(\{\textbf{Y}_k^{(t)}: k<j\}, \{\textbf{Y}_k^{(t-1)}: k>j\})$, the analyst first generates predictions for each tree based on the values $\textbf{Y}_{j}^{\textrm{obs}}$ in the corresponding leaf for that tree, and then uses the most commonly occurring majority level of among all predictions from all the trees. For a continuous $\textbf{Y}_{(j)}$, the analyst instead uses the average of all the predictions from all the trees. The iterative process again cycles through all the variables, for $T$ total iterations, and the values at the final iteration make up a completed dataset. A particularly important hyperparameter in random forests is the maximum number of trees $d$.

For our evaluations, we use the \texttt{mice} R package to implement both MICE-CART and MICE-RF, and retain the default hyperparameter setting in the package to mimic the common practice in real world applications. Specifically, we set the minimum number of observations in each terminal leaf to 5 and the pruning threshold to 0.0001 in MICE-CART. In MICE-RF, the maximum number of trees $d$ is set to be 10.

\subsection{Generative Adversarial Imputation Network (GAIN)}

GAIN \citep{yoon2018gain} is an imputation method based on GANs \citep{goodfellow2014gan}, which consist of a generator function ${\emph G}$ and a discriminator function ${\emph D}$. For any data matrix ${\bf Y} = ({\bf Y}_{\textrm{obs}}, {\bf Y}_{\textrm{mis}})$, we replace ${\bf Y}_{\textrm{mis}}$ with random noise, $Z_{ij}$, sampled from a uniform distribution. The generator ${\emph G}$ inputs this initialized data and a mask matrix ${\bf M}$, with $M_{ij} \in \{0,1\}$ indicating observed values of ${\bf Y}$, and outputs predicted values for both the observed data and missing data, $\hat{\bf Y}$. The discriminator ${\emph D}$ utilizes $\hat{\bf Y}$ = (${\bf Y}_{\textrm{obs}}$, $\hat{\bf Y}_{\textrm{mis}}$) and a hint matrix ${\bf H}$ of the same dimension to identify which values are observed or imputed by ${\emph G}$, which results in a predicted mask matrix $\hat{\bf M}$. The hint matrix, sampled from the Bernoulli distribution with $p$ equal to a ``hint rate'' hyperparameter, reveals to ${\emph D}$ partial information about ${\bf M}$ in order to help guide ${\emph G}$ to learn the underlying distribution of ${\bf Y}$.  

We first train ${\emph D}$ to minimize the loss function, ${L}_D({\bf M}, \hat{\bf M})$, for each mini-batch of size $n_i$: 
\begin{equation}
    {L}_D({\bf M}, \hat{\bf M}) = \sum_{i=1}^{n_i}\sum_{j = 1}^{J}M_{ij} \, \text{log}(\hat{M}_{ij}) + (1 - M_{ij}) \, \text{log}(1-\hat{M}_{ij}).
    \label{D_loss}
\end{equation}
Next, ${\emph G}$ is trained to minimize the loss function \eqref{G_loss}, which is composed of a generator loss, ${L}_G({\bf M}, \hat{\bf M})$, and a reconstruction loss, ${L}_M({\bf Y}, \hat{\bf Y}, {\bf M})$. The generator loss \eqref{generator_loss} is minimized when ${\emph D}$ incorrectly identifies imputed values as being observed. The reconstruction loss \eqref{reconstruction_loss} is minimized when the predicted values are similar to the observed values, and is weighted by the hyperparameter $\beta$:
\begin{align}
    {L}({\bf Y}, \hat{\bf Y}, {\bf M}, \hat{\bf M}) &= {L}_G({\bf M}, \hat{\bf M}) + \beta {L}_M({\bf Y}, \hat{\bf Y}, {\bf M}),
    \label{G_loss} \\
    {L}_G({\bf M}, \hat{\bf M}) &= \sum_{i=1}^{n_i}\sum_{j = 1}^{J}M_{ij} \, \text{log}(1 - \hat{M}_{ij}),
    \label{generator_loss}\\
    {L}_M({\bf Y}, \hat{\bf Y}, {\bf M}) &= \sum_{i=1}^{n_i}\sum_{j = 1}^{J}(1 - M_{ij}) \, L_{\textrm{rec}}(Y_{ij}, \hat{Y}_{ij}),
    \label{reconstruction_loss}
\end{align}
where 
\begin{equation}
    L_{\textrm{rec}}(Y_{ij}, \hat{Y}_{ij}) = 
        \begin{cases}
            (\hat{Y}_{ij} - Y_{ij})^2         &\text{if $Y_{ij}$ is continuous}\\
            -Y_{ij} \, \text{log}\hat{Y}_{ij}     &\text{if $Y_{ij}$ is categorical}.
        \end{cases}
    \label{rec}
\end{equation}
In our experiments, we model both ${\emph G}$ and ${\emph D}$ as fully-connected neural networks, each with three hidden layers, and $\theta$ hidden units per hidden layer. The hidden layer weights are initialized uniformly at random with the Xavier initialization method \citep{glorot2010}. We use leaky ReLU activation function \citep{maas2013rectifier} for each hidden layer, and a softmax activation function for the output layer for ${\emph G}$ in the case of categorical variables, or a sigmoid activation function in the case of numerical variables and for the output of ${\emph D}$. We facilitate this choice of output layer for numerical variables by transforming all continuous variables to be within range (0, 1) using
the MinMax normalization: $Y_{ij}^* = \{Y_{ij} - \text{min}(Y_{\cdot j})\}/\{\text{max}(Y_{\cdot j}) - \text{min}(Y_{\cdot j})\}$, where $\text{min}(Y_{\cdot j})$ and $\text{max}(Y_{\cdot j})$ are the minimum and maximum of variable $j$, respectively. After imputation, we transform each value back to its original scale.    We generate multiple imputations using several runs of the model with varying initial imputation of the missing values.

To implement GAIN in our evaluations, we use the same architecture as the one in \citet{yoon2018gain}. We set $\beta =100$, $\theta$ equal to the number of features of the input data, and tune the hint rate on a single simulation. Following the common practice in the GAN literature \citep{DBLP:journals/corr/BerthelotSM17, ham2020unbalanced}, we track the evolution of GAIN's generator and discriminator losses, and manually tune the hint rate so that the two losses are qualitatively similar. Specifically, we first coarsely select the hint rate among \{0.1, 0.2, 0.3, 0.4, 0.5, 0.6, 0.7, 0.8, 0.9 \}. Then we determine the final value by an additional fine tuning step. In the MAR scenario, for example, after observing that the optimal value is in the range (0.1, 0.2), we perform a search among \{ 0.11, 0.12, 0.13, 0.14, 0.15, 0.16, 0.17, 0.18, 0.19 \}. Finally, we set the optimal hint rate for MCAR and MAR scenarios to be 0.3 and 0.13, respectively. We train the networks for 200 epochs using stochastic gradient descent (SGD) and mini-batches of size 512
to learn the parameter weights. We use the Adam optimizer to adapt the learning rate, with an initial rate of 0.001 \citep{kingma2014adam}. 


\subsection{Multiple Imputation using Denoising Autoencoders (MIDA)}

MIDA \citep{gondara2017mida,lu2020multiple} extends a class of neural networks, denoising autoencoders, for MI. An autoencoder is a neural network model trained to learn the identity function of the input data. Denoising autoencoders intentionally corrupt the input data in order to prevent the networks from learning the identity function, but rather a useful low-dimensional representation of the input data. The MIDA architecture consists of an encoder and decoder, each modeled as a fully-connected neural network with three hidden layers, with $\theta$ hidden units per hidden layer. We first perform an initial imputation on missing values using the mean for continuous variables and the most frequent label for categorical variables, which results in a completed data ${\bf Y}_0$. The encoder inputs ${\bf Y}_0$, and corrupts the input data by randomly dropping out half of the variables. The corrupted input data is mapped to a higher dimensional representation by adding $\Theta$ hidden units to each successive hidden layer of the encoder. The decoder receives output from the encoder, and symmetrically scales the encoding back to the original input dimension. All hidden layers use a hyperbolic tangent (tanh) activation function, while the output layer of the decoder uses a softmax (sigmoid) activation function in the case of categorical (numerical) variables. Multiple imputations are generated by using multiple runs with the hidden layer weights initialized as a Gaussian random variable.

Following \citet{lu2020multiple}, we train MIDA in two phases: a primary phase and fine-tuning phase. In the primary phase, we feed the initially imputed data to MIDA and train for $N_{\textrm{prime}}$ epochs. In the fine-tuning phase, MIDA is trained for $N_{\textrm{tune}}$ epochs on the output in the primary phase, and produces the outcome. The loss function is used in both phases and closely resembles the reconstruction loss in GAIN:
\begin{equation}
    L({Y}_{{ij}_0}, \hat{Y}_{ij}, M_{ij}) = 
    \begin{cases}
        (1 - M_{ij}) ({Y}_{{ij}_0} - \hat{Y}_{ij})^2 &\text{if $Y_{ij}$ is continuous}\\
        -(1 - M_{ij}){Y}_{{ij}_0} \, \text{log}\hat{\bf Y}_{{ij}}    &\text{if $Y_{ij}$ is categorical}.
    \end{cases}
    \label{MIDA_loss}
\end{equation}
To implement MIDA in our evaluations, we use the same architecture and tune the hyperparameters in a single simulation as in \citeauthor{lu2020multiple}. We plot the evolution of loss function $L$, and select the number of additional units $\Theta$ among \{1, 2, 3, 4, 5, 6, 7 ,8, 9, 10 \} to reduce the loss. In our experiments, we set $\theta$ equal to the number of features of the input data and add $\Theta=7$ hidden units to each of the three hidden layers of the encoder. We train the model for $N_{\textrm{prime}}=100$ epochs in the primary phase and $N_{\textrm{tune}}=2$ epochs in the fine-tuning phase. Similar as in GAIN, we learn the model parameters using SGD with mini-batches of size 512, and use the Adam optimizer to adapt the learning rate with the initial rate being 0.001. 


\section{Simulation-based evaluation of imputation methods} \label{sec:framework}

Methods for missing data imputation are usually evaluated via real-data based simulations \citep{van2018flexible}. Namely, one creates missing values from a complete dataset according to a missing data mechanism \citep{little2014}, imputes the missing values by a specific method, and then compares these imputed values with the original ``true'' values based on some metrics. 


We first give a quick review of Rubin's MI combination rules. Let $Q$ be the target estimand in the population, and $q^{(l)}$ and $u^{(l)}$ be the point and variance estimate of $Q$ based on the $l$th imputed dataset, respectively. The MI point estimate of $Q$ is $\bar{q}_L = \sum_{l=1}^{L}q^{(l)}/L$, and the corresponding estimate of the variance is equal to $T_L = (1 + 1/L) b_L + \bar{u}_L$, where $b_L = \sum_{l=1}^L (q^{(l)} - \bar{q}_L)^2/(L-1)$, and $\bar{u}_L = \sum_{l=1}^L u^{(l)}/L$. The confidence interval of $Q$ is constructed using $(\bar{q}_L - Q) \sim t_{\nu}(0, T_L)$, where $t_{v}$ is a $t$-distribution with $\nu = (L-1)(1 + \bar{u}_L/ [(1+1/L) b_L])^2$ degrees of freedom.

The first step in our simulation-based evaluation procedure is choosing a dataset with all values observed, which is taken as the ``population.'' We then choose a set of target estimands $Q$ and compute their values from this population data, which are taken as the ``ground truth.'' The estimands are usually summary statistics of the variables or parameters in a down-stream analysis model, e.g., a coefficient in a regression model \citep{tang2005comparison, huque2018comparison}. Second, we randomly draw without replacement $H$ samples of size $n$ from the population data, and in each of sample ($h=1,...,H$) create missing data according to a specific missing data mechanism and pre-fixed proportion of missingness. Third, for each simulated sample with missing data, we create $L$ imputed datasets using the imputation method under consideration and construct the point and interval estimate of each estimand using Rubin's rules. Lastly, we compute performance metrics of each estimand from the quantities obtained in the previous step.    

In the empirical application, we select a large complete subsample from the American Community Survey (ACS) --- a national survey that bears the hallmarks of many big survey data --- as our population.  Since discrete variables are prevalent in the ACS, as well as in most survey data, we focus on the marginal probabilities of binary and categorical variables; e.g., a categorical variable with $K$ categories has $K-1$ estimands. To evaluate how well the imputation methods preserve the multivariate distributional properties, similar to \citet{akande2017empirical}, we also consider the bivariate probabilities of all two-way combinations of categories in binary and categorical variables. Another useful metric is the finite-sample pairwise correlations between continuous variables.   For continuous variables, the common estimands are mean, median or variance. To facilitate meaningful comparisons of the results between the categorical and continuous variables, we propose to discretize each continuous variable into $K$ categories based on the sample quantiles. We then evaluate these binned continuous variables as categorical variables based on the aforementioned estimands of marginal and bivariate probabilities.

For each estimand $Q$, we consider three metrics. The first metric focuses on bias. To accommodate close-to-zero estimands that are prevalent in probabilities of categorical variables, we consider the absolute standardized bias (ASB) of each estimand $Q$:
\begin{equation}
    \text{ASB} = {\sum_{h=1}^{H}|\bar{q}^{(h)}_L - Q}|/{(H\cdot Q)}, \label{eq:ASB}
\end{equation}
where $\bar{q}^{(h)}_L$ is the MI point estimate of $Q$ in simulation $h$.

The second metric is the relative mean squared error (Rel.MSE), which is the ratio between the MSE of estimating $Q$ from the imputed data and that from the sampled data before introducing the missing data:
\begin{equation}
    \text{Rel.MSE} = \frac{\sum_{h=1}^{H}(\bar{q}^{(h)}_L - Q)^2}
                          {\sum_{h=1}^{H}(\widetilde{Q}^{(h)} - Q)^2}, \label{eq:relMSE}
\end{equation}
where $\bar{q}^{(h)}_L$ is defined earlier, and $\widetilde{Q}^{(h)}$ is the prototype estimator of $Q$, i.e. the point estimate from the complete sampled data in simulation $h$.

The third metric is coverage rate, which is the proportion of the $\alpha\%$ (e.g. 95\%) confidence intervals, denoted by $\mbox{CI}_h^{\alpha}$ ($h=1,...,H$), in the $H$ simulations that contain the true $Q$: 
\begin{equation}
    \text{Coverage} = \sum_{h=1}^{H}\one\{Q\in\mbox{CI}_h^{\alpha}\}/H. \label{eq:coverage}
\end{equation}

We recommend conducting a large number of simulations (e.g. $H\geq 100$) to obtain reliable estimates of MSE and coverage. This would not be a problem for deep learning algorithms, which can be typically completed in seconds even with large sample sizes. However, it can be computationally prohibitive for the MICE algorithms when each of the simulated data is large (e.g. $n=100,000$ in some of our simulations). In the situation that one has to rely on only a few or even a single simulation for evaluation, we propose a modified metric of bias. 
Specifically, for each categorical variable or binned continuous variable $j$ , we define the weighted absolute bias (WAB) as the sum of the absolute bias weighted by the true marginal probability in each category: 
\begin{equation}
    \text{Weighted absolute bias} = \sum_{k = 1}^K Q_{jk}\left|\bar{q}^{(h)}_{jk} - Q_{jk}\right|,
    \label{eq: weighted_bias}
\end{equation}
where $K$ is the total number of categories, $Q_{jk}$ is the population marginal probability of category $k$ in variable $j$, and $\bar{q}^{(h)}_{jk}$ is its corresponding point estimate in simulation $h$. We can also average the weighted absolute bias over a number of repeatedly simulated samples.  

The above procedure and metrics differ from the common practice in the machine learning literature. For example, many machine learning papers on missing data imputation conduct simulations on benchmark datasets, but these data often have vastly different structure and features from survey data and thus are less informative for the goal of this paper. One such dataset is the Breast Cancer dataset in the UCI Machine Learning Repository \citep{Dua2019}, which has only 569 sample units and no categorical variables. Also, these simulations are  usually based on randomly creating missing values of a single dataset repeatedly rather than on drawing repeated samples from a population, and thus fails to account for the sampling mechanism. Moreover, these evaluations often use metrics focusing on accuracy of individual predictions rather than distributional features. Specifically, the most commonly used metrics are the root mean squared error (RMSE) and accuracy \citep{gondara2017mida, yoon2018gain,lu2020multiple}. Both metrics can be defined in an overall or variable-specific fashion, but the machine learning literature usually focuses on the overall version. The overall RMSE is defined as  
\begin{equation}
    \label{eq:rmse}
    \text{RMSE} = \sqrt{\frac{\sum^{n}_{i=1}\sum_{j}M_{ij}(\hat{Y}_{ij} - Y_{ij})^2}
                       {\sum^{n}_{i=1}\sum_{j}M_{ij}}},
\end{equation}
where $Y_{ij}$ is the value of continuous variable $j$ for individual $i$ in the complete data before introducing missing data, and $\hat{Y}_{ij}$ is the corresponding imputed value. For non-missing values (i.e. $M_{ij} = 1$), $Y_{ij} = \hat{Y}_{ij}$. The (overall) accuracy is defined for categorical variables, namely it is the proportion of the imputed values being equal to the corresponding original ``true'' value:
\begin{equation}
    \label{eq:accuracy}
    \text{Accuracy} = \frac{\sum^{n}_{i=1}\sum_{j \in S_{cat}}M_{ij}\mathbbm{1}(\hat{Y}_{ij} = Y_{ij})}{\sum^{n}_{i=1}\sum_{j \in S_{cat}}M_{ij}},
\end{equation}
where $S_{cat}$ is the set of categorical variables. 

A number of caveats are in order for the RMSE and accuracy metrics. First, they are usually computed on a single imputed sample as an overall measure of an imputation method, but this ignores the uncertainty of imputations. Second, both RMSE and accuracy are single value summaries and do not capture the multivariate distributional feature of data. Third, RMSE does not adjust for the different scale of variables and can be be easily dominated by a few outliers; also, it is often computed without differentiating between continuous and categorical variables. Lastly, when there are multiple ($L$) imputed data, a common way is to use the mean of the $L$ imputed value as $\hat{Y}_{ij}$ in \eqref{eq:rmse}, but the statistical meaning of the resulting metrics is opaque. This is particularly problematic for categorical variables. For these reasons, we warn against using the overall RMSE and accuracy as the only metrics for comparing imputation methods, and one should exercise caution when interpreting them.

\section{Evaluation based on ACS} \label{sec:evaluation}
In this section, we evaluate the four imputation methods described in Section  \ref{sec:methods} following the procedure and metrics described in Section \ref{sec:framework}. For simplicity, in the following discussions we use CART and RF to denote MICE-CART and MICE-RF, respectively. 

\subsection{The ``population'' data}
We use the one-year Public Use Microdata Sample from the 2018 ACS to construct our population. The 2018 ACS data contains both household-level variables --- for example, whether or not a house is owned or rented --- and individual-level variables --- for example, age, income and sex of the individuals within each household. Since individuals nested within a household are often dependent, and the imputation methods we evaluate generally assume independence across all observations, we set our unit of observation at the household-level, where independence is more likely to hold. We first remove units corresponding to vacant houses. Next, we delete units with any missing values, so that we only keep the complete cases. Within each household, we also retain individual-level data corresponding only to the household head and merge them with the household-level variables, resulting in a rich set of variables with potentially complex joint relationships.

It is often challenging to generate plausible imputations for ordinal variables with many levels when there is very low mass at the highest levels, as is the case for some variables in the ACS data. Following \citet{LiEtAl2014:JCGS}, we treat ordinal variables with more than 10 levels as continuous variables. We also follow the approach in \citet{akande2017empirical} to exclude binary variables where the marginal probabilities violate $np > 10$ or $n(1-p)>10$; this eliminates estimands where the central limit theorem is not likely to hold. For each categorical variable with more than two levels but less than 10 levels where this might also be a problem, we merge the levels with a small number of observations in the population data. For example, for the household language variable, we recode the levels from five to three (English, Spanish, and other), because the probability of speaking neither English nor Spanish in the full population is less than 8.8\%.

The final population data contains 1,257,501 units, with 18 binary variables, 20 categorical variables with 3 to 9 levels, and 8 continuous variables. We describe the variables in more detail in the supplementary material. We compute the population values of the estimands $Q$ described in Section \ref{sec:framework}, including all marginal and bivariate probabilities of discrete and binned continuous variables. We vary the size of the simulated samples from 10,000 to 100,000, and simulate missing data according to either missing completely at random (MCAR) or missing at random (MAR) mechanisms in each of these scenarios.

\subsection{Simulations with n=10,000} \label{sec:evaluation:n10000}
We first randomly draw $H=100$ samples of size $n = 10,000$, and set $30\%$ of each sample to be missing under either MCAR or MAR. CART or RF takes around 2.8 and 9.2 hours, respectively, to create $L=10$ imputed datasets with default parameters on a standard desktop computer with a single central processing unit (CPU). The deep learning methods are much faster because they leverage GPU computing power when implemented on the GPU-enabled TensorFlow software framework \citep{tensorflow2015-whitepaper}. GAIN takes roughly 1.5 minutes and MIDA takes roughly 4 minutes to create $L=10$ completed datasets using a GeForce GTX 1660 Ti GPU. Note that it is infeasible to manually tune the hyperparameter in each of the 100 simulations in each scenario for the deep learning models. So for each scenario, we have randomly selected one simulation, and tune the hyperparameters using the procedure described in Section \ref{sec:methods}. We then apply these selected hyperparameters to all simulations.

\subsubsection{MCAR scenario}
To create the MCAR scenario, we randomly set 30\% of the values of each variable to be missing independently. Table \ref{results:asb_relmse_mcar10000} displays the distributions of the estimated ASB and relative MSE of all the marginal and bivariate probabilities in the imputed data by the four imputation methods.
\begin{table}
\caption{\label{results:asb_relmse_mcar10000}Distributions of absolute standardized bias $(\times 100)$ and relative mean squared error of all marginal and bivariate probabilities based on the imputations by the four MI methods, when $n=10,000$ and 30\% values MCAR. ``Cat.'' means categorical variables and ``B.Cont.'' means binned continuous variables.}
\centering
\begin{tabular}{r r cccc c cccc}
\toprule
\multicolumn{2}{r}{\multirow{2}{*}{Quantiles}} & \multicolumn{4}{c}{Marginal} && \multicolumn{4}{c}{Bivariate}  \\
\cmidrule{3-6} \cmidrule{8-11}
&  &      CART &    RF &  GAIN &  MIDA & &      CART &    RF &  GAIN &  MIDA \\
\midrule
\vspace{-6pt} \\
\multicolumn{11}{c}{\underline{ASB $(\times 100)$}}\\
                &10\% &     0.05 &  0.47 &  0.76 &  0.98 &&      0.15 &  1.14 &  1.21 &  1.54      \\
                &25\% &     0.13 &  1.25 &  1.48 &  2.22 &&      0.40 &  2.83 &  3.08 &  3.93      \\
Cat.            &50\% &     0.27 &  2.80 &  3.22 &  4.69 &&      1.05 &  6.74 &  7.14 &  8.47      \\
                &75\% &     0.64 &  5.86 &  7.18 &  8.86 &&      2.51 & 13.59 & 17.03 & 15.23      \\
                &90\% &     1.14 & 10.01 & 19.55 & 14.41 &&      5.34 & 22.33 & 26.92 & 21.90      \\
\midrule
                &10\% &     0.06 & 0.24 &  7.25 &  2.73 &&      0.19 &  1.30 &  6.05 &  4.80\\
                &25\% &     0.10 & 1.05 & 12.86 &  8.36 &&      0.43 &  3.24 & 17.61 & 12.01\\
B.Cont.         &50\% &     0.21 & 3.59 & 27.30 & 18.51 &&      1.02 &  6.61 & 34.29 & 24.07\\
                &75\% &     0.43 & 5.43 & 30.21 & 26.84 &&      1.90 & 11.76 & 49.38 & 39.54\\
                &90\% &     0.81 & 8.49 & 46.41 & 31.36 &&      3.42 & 20.79 & 90.90 & 64.65\\
\midrule
\vspace{-6pt} \\
\multicolumn{11}{c}{\underline{Rel.MSE}}\\
                &10\% &     1.05 &  1.67 &  2.50 &  3.38 &&      0.96 &  1.11 &  2.75 &  2.98      \\
                &25\% &     1.16 &  2.40 &  4.97 &  9.03 &&      1.08 &  1.61 &  4.33 &  4.75      \\
Cat.            &50\% &     1.37 &  5.99 & 10.37 & 14.89 &&      1.25 &  3.35 &  7.40 &  8.16      \\
                &75\% &     1.49 & 10.25 & 27.73 & 26.16 &&      1.48 &  9.07 & 14.87 & 15.80      \\
                &90\% &     1.62 & 16.22 & 97.33 & 40.16 &&      1.89 & 23.91 & 36.37 & 27.92      \\
\midrule
                &10\% &     1.19 &  1.50 &  44.06 &   4.35 &&      0.82 &  0.86 &   7.40 &   2.05\\
                &25\% &     1.30 &  1.77 &  74.42 &  13.82 &&      0.92 &  1.11 &  14.80 &   4.90\\
B.Cont.         &50\% &     1.44 &  3.31 & 139.24 &  72.57 &&      1.07 &  1.90 &  32.26 &  13.76\\
                &75\% &     1.55 &  6.71 & 284.00 & 150.35 &&      1.26 &  4.09 &  88.78 &  47.56\\
                &90\% &     1.64 & 19.69 & 603.38 & 451.44 &&      1.54 & 10.80 & 282.29 & 127.15\\
\bottomrule
    \end{tabular}
\end{table}

Overall, for the estimands of marginal and bivariate probabilities of the categorical and binned continuous variables, MICE with CART significantly outperforms all other three methods, with consistently yielding the smallest ASB and relative MSE.  RF is the second best, also consistently outperforming the deep learning methods. The advantage of the MICE algorithms is particularly pronounced in the upper (e.g. 75\% and 90\%) quantiles, indicating that GAIN and MIDA imputations have large variations over repeated samples and variables. Indeed, MIDA and GAIN lead to ultra long tails in estimating the summary statistics of the variables. For example, for bivariate probabilities of binned continuous variables, the 90\% percentile of the ASB from MIDA and GAIN is approximately 20 and 27 times, respectively, of that from CART. The discrepancy is even bigger for relative MSE. There is no consistent pattern in comparing MIDA and GAIN. Specifically, for continuous variables, MIDA generally outperforms GAIN, but the difference is small except for the upper percentiles, where GAIN tends to produce very large bias and relative MSE. For categorical variables, GAIN outperforms MIDA half of the time, but again leads to the largest variation in imputations across the variables.  Moreover, an interesting and somewhat surprising observation is that MICE with CART consistently outperforms RF---sometimes by a large magnitude---regardless of the choice of estimand or metric.

All methods generally yield less biased estimates (i.e. smaller ASB) of the marginal probabilities than the bivariate probabilities. This illustrates preserving multivariate distributional features is more challenging than univariate ones. The advantage of CART over the other methods is comparatively larger when estimating bivariate estimands than univariate ones. Interestingly, the relative MSE tends to be higher for the marginal probabilities than the bivariate probabilities. This is likely due to the fact that the denominator in the definition of relative MSE in (\ref{eq:relMSE}) is the MSE from the sampled data before introducing missing data, which tends to be smaller for marginal probabilities than bivariate probabilities. CART yields MSEs that are very close to the corresponding MSEs from the sampled data before introducing missing data; i.e., the relative MSE is close to 1. On the contrary, both deep learning methods, and GAIN in particular, can result in exceedingly large relative MSE for many estimands. 


%

\begin{figure}
 \centering
 \includegraphics[width=0.85\textwidth, angle=0]{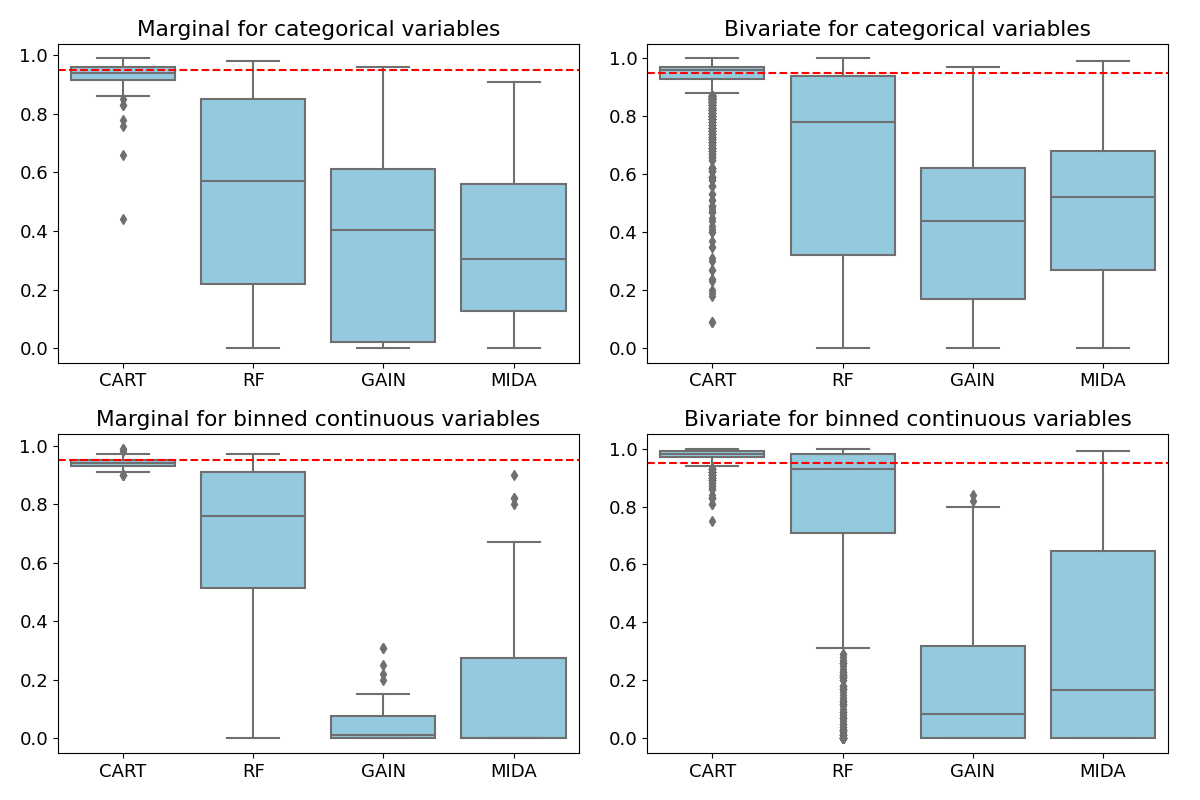}
 \caption{Coverage rate of the 95\% confidence interval for all marginal and bivariate probabilities obtained from the four imputation methods in the simulations with $n=10,000$ and 30\% values MCAR. The red dashed line is 0.95.}
   \label{figure:house_cov}
\end{figure}
Figures \ref{figure:house_cov} displays the estimated coverage rates of the $95\%$ confidence intervals for the marginal and bivariate probabilities. The patterns on coverage between different methods is similar to those on bias and MSE. Specifically, CART tends to result in coverage rates that are close to the nominal $95\%$ level, with the median consistently being around 95\% and tight interquartile range. In contrast, RF, GAIN and MIDA all result in coverage rates that are much farther off from the nominal $95\%$ level. For example, the median coverage rates under both GAIN and MIDA are all under 0.60, and are even less than 0.30 for continuous variables. A closer look into the prediction accuracy of each variable reveals that GAIN and MIDA tend to generate imputations that are biased toward the most frequent levels, and GAIN in particular generally produces narrower intervals than the other methods. This once again provides evidence of significant bias under the deep learning methods. All methods tend to result in higher median coverage rates for the bivariate probabilities than the marginal probabilities, although the left tails are generally longer for the former than the latter.


\subsubsection{MAR scenario}

We also consider a MAR scenario, which is more plausible than MCAR in practice. We set six variables --- age, gender, marital status, race, educational attainment and class of worker --- to be fully observed. It would be cumbersome to specify different MAR mechanism for each of the remaining 40 variables, so we randomly divide them into three groups, consisting of 10, 15, and 15 variables. We then specify a separate nonresponse model by which to generate the missing data for the variables in each group. Specifically, we postulate a logistic model per group, conditional on the fully observed six variables, based on which we then generate binary missing data indicators for each variable in that group. This process results in approximately 30\% missing rate for each of the 40 variables. We describe the models in more detail in the supplementary material.

\begin{table}
\caption{\label{results:asb_relmse_mcar10000_mar}Distributions of absolute standardized bias $(\times 100)$ and relative mean squared error for all methods, when $n=10,000$ and 30\% values MAR, over all possible marginal and bivariate probabilities. ``Cat.'' means categorical variables and ``B.Cont.'' means binned continuous variables.}
\centering
\begin{tabular}{r r  cccc c cccc}
\toprule
\multicolumn{2}{r}{\multirow{2}{*}{Quantiles}} & \multicolumn{4}{c}{Marginal} && \multicolumn{4}{c}{Bivariate}  \\
\cmidrule{3-6} \cmidrule{8-11}
&  &      CART &    RF &  GAIN &  MIDA & &      CART &    RF &  GAIN &  MIDA \\
\midrule
\vspace{-6pt} \\
\multicolumn{11}{c}{\underline{ASB $(\times 100)$}}\\
                &10\% &     0.05 &  0.13 &  0.15 &  0.14 &&      0.15 &  0.71 &  0.76 &  0.89      \\
                &25\% &     0.11 &  0.44 &  0.62 &  0.61 &&      0.40 &  2.23 &  2.55 &  3.20      \\
Cat.     &50\% &     0.29 &  2.13 &  3.05 &  4.55 &&      1.08 &  6.06 &  6.85 &  8.14      \\
                &75\% &     1.04 &  4.98 &  6.63 & 10.22 &&      2.49 & 13.43 & 16.78 & 16.19      \\
                &90\% &     1.80 & 10.49 & 18.91 & 17.00 &&      5.68 & 24.06 & 28.04 & 25.36     \\
\midrule
                &10\% &     0.07 &  0.29 &  0.33 &  0.33 &&      0.27 &  1.17 & 10.87 &  6.18\\
                &25\% &     0.17 &  1.07 &  9.64 &  3.13 &&      0.69 &  3.49 & 23.67 & 16.26\\
B.Cont.     &50\% &     0.67 &  3.14 & 32.86 & 23.85 &&      1.58 &  7.83 & 38.52 & 31.17\\
                &75\% &     1.20 &  6.95 & 39.57 & 36.09 &&      3.40 & 15.20 & 53.59 & 47.34\\
                &90\% &     3.40 & 12.39 & 63.45 & 41.99 &&      5.94 & 25.16 & 97.47 & 85.44\\
\midrule
\vspace{-6pt} \\
\multicolumn{11}{c}{\underline{Rel.MSE}}\\
                &10\% &     1.00 &  1.00 &   1.00 &  1.00 &&      0.97 &  1.00 &  1.53 &  1.93      \\
                &25\% &     1.08 &  1.82 &   2.56 &  4.75 &&      1.04 &  1.39 &  3.78 &  4.03      \\
Cat.     &50\% &     1.33 &  4.33 &  19.03 & 15.13 &&      1.25 &  3.00 & 10.42 &  8.38      \\
                &75\% &     1.72 & 13.08 &  55.07 & 33.36 &&      1.59 &  9.56 & 27.45 & 16.95      \\
                &90\% &     2.27 & 18.70 & 101.91 & 48.44 &&      2.23 & 27.44 & 64.01 & 32.85     \\
\midrule
                &10\% &     1.00 &  1.00 &    1.00 &   1.00 &&      0.88 &  0.90 &  11.19 &   2.96\\
                &25\% &     1.38 &  1.83 &   90.98 &   8.49 &&      1.00 &  1.16 &  20.15 &   6.87\\
B.Cont.     &50\% &     1.70 &  4.57 &  207.58 &  96.08 &&      1.18 &  2.29 &  45.25 &  21.33\\
                &75\% &     2.12 & 11.47 &  692.67 & 239.69 &&      1.50 &  6.95 & 125.39 &  70.90\\
                &90\% &     3.12 & 50.56 & 1342.23 & 806.43 &&      2.12 & 18.07 & 459.78 & 205.14\\
\bottomrule
    \end{tabular}
\end{table}
Table \ref{results:asb_relmse_mcar10000_mar} displays the distributions of the ASB and relative MSE of all the marginal and bivariate probabilities from the four methods. All methods yield larger ASB and relative MSE under the MAR scenario than the previous MCAR scenario. This is expected because MAR is a stronger assumption than MCAR that requires conditioning on more information. Nonetheless, the overall patterns of relative performance between the methods remain the same as those under MCAR. Specifically, CART once again produces estimates with the least ASB and relative MSE --- by an even larger margin then under MCAR --- among the four methods, followed by RF, and then MIDA and GAIN. One notable observation is the deteriorating performance of the deep learning methods, particularly GAIN, in imputing continuous variables, sometimes resulting in several hundreds fold of relative MSE than CART. This indicates the huge uncertainties associated with GAIN in imputing continuous variables.   

\begin{figure}
 \centering
 \includegraphics[width=0.85\textwidth, angle=0]{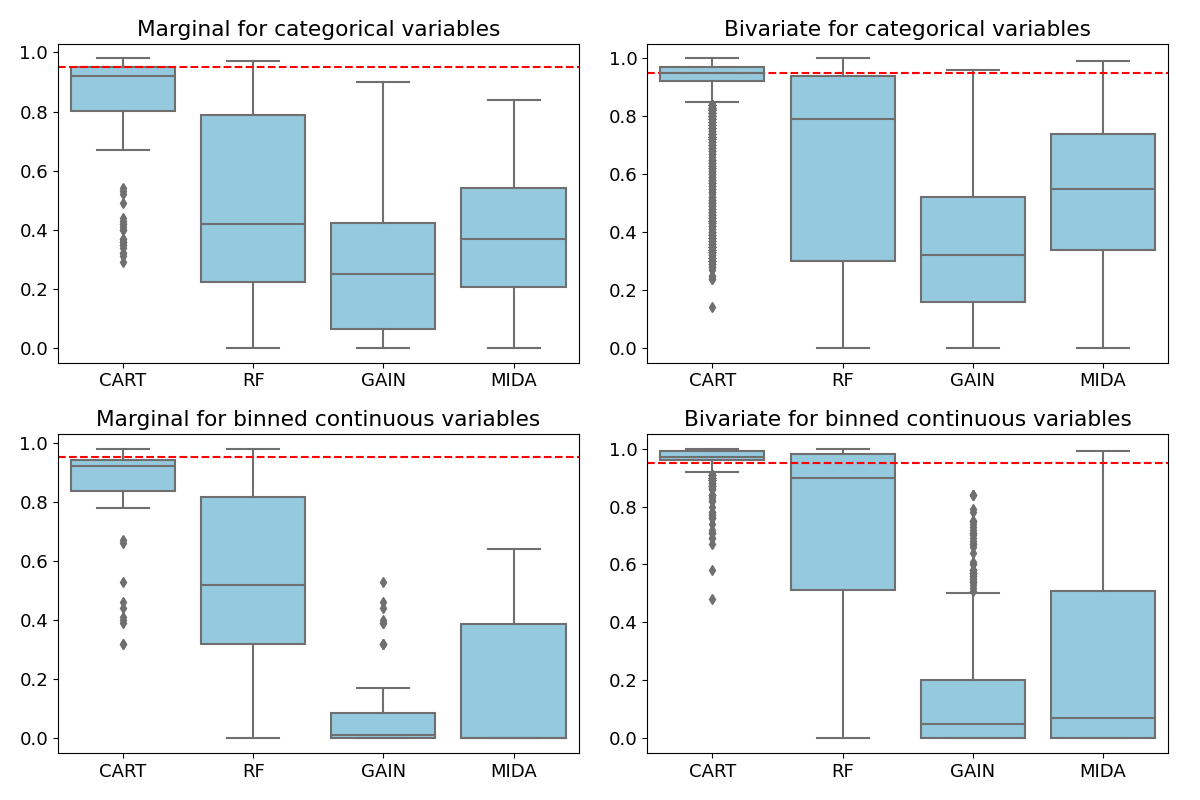}
 \caption{Coverage rates of the 95\% confidence intervals for all marginal and bivariate probabilities obtained from four methods in the simulations with $n=10,000$ and 30\% values MAR. The red dashed line is 0.95.}
   \label{figure:house_cov_cat_mar}
 \end{figure}
Figures \ref{figure:house_cov_cat_mar} displays the estimated coverage rates of the $95\%$ confidence intervals for the marginal and bivariate probabilities, under each method. Similar as the case of bias and MSE, all methods generally result in lower coverage rates under MAR than MCAR, with visibly longer left tails in some cases, but the overall patterns comparing between the methods remain the same. Specifically, CART still tends to result in coverage rates that are above 90\%, while the other three methods have consistently lower coverage rate. In particular, both GAIN and MIDA result in extremely low---below 7\%-- median coverage rates for continuous variables. This is closely related to the previous observation of the large uncertainty of the deep learning methods in imputing continuous variables.

Finally, to illustrate that evaluating only the overall RMSE and accuracy metrics may be misleading, we display the mean and empirical standard errors of the overall RMSE and accuracy over the 100 simulations in Table \ref{table:rmseacc}, where MCAR is in the top panel and MAR is in the bottom panel. Under both missing data mechanisms, for the continuous variables, MIDA leads to the smallest overall RMSE, followed by CART, and with RF and GAIN being last. For the categorical variables, CART and GAIN lead to the highest overall accuracy, with MIDA being closely behind and RF last. These patterns, not surprisingly, differ from those reported earlier based on marginal and bivariate probabilities and different metrics. As discussed in Section \ref{sec:framework}, overall RMSE and accuracy do not capture the distributional features of multivariate data or the repeated sampling properties of the imputation methods.   
\begin{table}
\caption{\label{table:rmseacc} Overall RMSE on continuous variables and overall accuracy on categorical variables averaged over 100 simulations, with the empirical standard errors in the parenthesis. The top panel is under MCAR and the bottom panel is under MAR, all with 30\% missing data.} 
\centering
\begin{tabular}{llrrrr}
\toprule
Mechanism & Metric&     CART &     RF &   GAIN &   MIDA \\
\midrule
&RMSE &      0.128 (0.002) & 0.159 (0.003) &        0.161 (0.008) &       0.112  (0.002)\\
MCAR &Accuracy &   0.785 (0.001) & 0.658 (0.003) &        0.782 (0.002) &       0.752 (0.004) \\
\hline
&RMSE &       0.130 (0.003)  & 0.154 (0.004) &       0.145 (0.009) &       0.110 (0.002)\\
MAR&Accuracy &   0.819 (0.001)  & 0.704 (0.003) &       0.820 (0.002) &       0.780 (0.007) \\
\bottomrule
\end{tabular}
\end{table}

\subsection{Simulations with n=100,000 and 30\% MCAR} 

Deep learning models usually require a large sample size to train. Therefore, to give MIDA and GAIN a more favorable setting as well as to investigate the sensitivity of our results to variations in sample size, we generate a simulation scenario of $H=10$ samples with $n = 100,000$ under MCAR. That is, we randomly set 30\% of the values of each variable to be missing independently. Here we only generate 10 simulations due to the huge computational cost of MICE for samples with this size. In this scenario, we omit RF because the previous results in Section \ref{sec:evaluation:n10000} have shown that RF is consistently inferior to CART in terms of performance and computation. We use CART, GAIN, and MIDA to create $L = 10$ completed datasets.

Because it usually requires a much larger number of simulations to reliably calculate MSE and coverage, here we focus on the weighted absolute bias metric \eqref{eq: weighted_bias}. Table \ref{table:wabias_100k} displays the distributions of the estimated weighted absolute bias, averaged over 10 simulations, of the marginal probabilities of the categorical and binned continuous variables. Overall, the patterns comparing between the four methods remain consistent with those observed in Section \ref{sec:evaluation:n10000}. Specifically, CART again results in the smallest weighted absolute difference in both categorical and continuous variables, and the advantage is particularly pronounced with continuous variables. For example, for categorical variables, MIDA and GAIN result in a median of weighted absolute bias at least 9 and 11 times, respectively, larger than CART. The advantage of CART grows to about 30 and 60 times over MIDA and GAIN, respectively, for continuous variables. Moreover, CART performs robustly across variables, evident from the small variation in the weighted absolute bias, e.g. 0.07 for 10\% percentile and 0.33 for 90\% percentile among the categorical variables. In contrast, both deep learning models result in much larger variation across variables; e.g., 0.57 for 10\% percentile and 2.92 for 90\% percentile among the categorical variables under MIDA, and even larger for GAIN. In summary, other than computational time, MICE with CART significantly outperforms MIDA and GAIN in terms of bias and variance regardless of the sample size.

\begin{table}
\caption{\label{table:wabias_100k} Distributions of the weighted absolute bias $(\times 100)$ averaged over 10 simulated samples, each with $n=100,000$ and 30\% values MCAR.}
\centering
\begin{tabular}{r ccc c ccc}
\toprule
\multirow{2}{*}{Quantiles } & \multicolumn{3}{c}{Categorical} && \multicolumn{3}{c}{Binned Continuous}  \\
\cmidrule{2-4} \cmidrule{6-8}
 &     CART &  GAIN &   MIDA  &&     CART &  GAIN &   MIDA \\
\midrule
10\% &     0.07 & 0.43 & 0.57  &&     0.10 & 5.52 & 1.98\\
25\% &     0.11 & 1.11 & 1.02  &&     0.11 & 6.65 & 2.78\\
50\% &     0.15 & 1.74 & 1.40  &&     0.12 & 7.36 & 4.04\\
75\% &     0.24 & 3.77 & 2.07  &&     0.13 & 9.40 & 6.50\\
90\% &     0.33 & 4.63 & 2.92  &&     0.15 & 11.31 & 7.72\\
\bottomrule
\end{tabular}
\end{table}

\subsection{Role of hyperparameters in tree-based MICE}
\begin{figure}[ht]
 \centering
 \includegraphics[width=0.9\textwidth, angle=0]{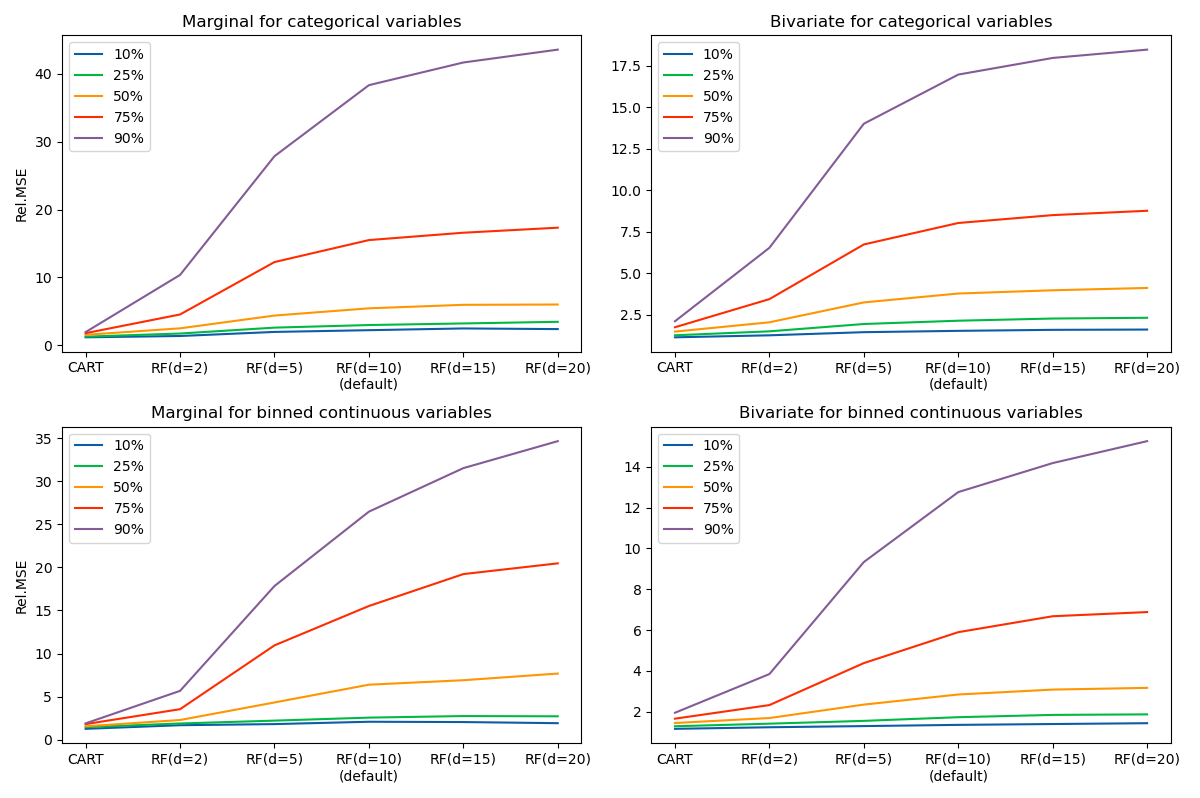}
 \caption{Quantiles of the relative mean squared error over all marginal and bivariate probabilities of categorical variables, under CART and RF with various number of trees, for a simulation sample with $n=10,000$ and $30\%$ values MCAR.}
   \label{figure:rf-diagnostics-relmse}
\end{figure}

The pattern that CART outperforms RF is surprising, because the common knowledge is that ensemble methods are usually superior to single tree methods. But the same pattern was also observed in another recent study \citep{wongkamthong2021comparative}. We investigate the role of the key hyperparameter in RF---the maximum number of trees $d$---in the simulations.  We randomly selected a simulated data of size $n=10,000$ and 30\% of entries being MCAR. We use the \texttt{mice} package to fit RF with different number of trees: $d=2, 5, 10, 15, 20$, where $d=10$ is the default setting. The relative MSE of the imputed categorical variables fitted using each $d$ value, as well as that using CART, is shown as trajectories in Figure \ref{figure:rf-diagnostics-relmse}, which reveals a consistent pattern: the upper quantiles ---particularly those above 50\%---of the relative MSE deteriorates rapidly as the maximum number of trees in RF increases, while the lower quantiles, e.g., 10\%, 25\%, remain stable. We found a similar pattern with the standardized bias metric and continuous variables, and thus the results are omitted here. This suggests that larger number of trees in RF---at least as implemented in the \texttt{mice} package---leads to much longer tail in the distribution of the bias and MSEs. This is likely due to overfitting. We cannot exclude the possibility that a more customized hyperparameter tuning of RF may outperform CART in some applications. However, such case-specific fine-tuning of the MICE algorithm is generally not available for the vast majority of MI consumers who relies on the default setting of popular packages like \texttt{mice}.
 
\section{Evaluation based on ``benchmark'' datasets} \label{sec:benchmark}

To verify the evaluations in the GAIN and MIDA papers \citep{gondara2017mida, yoon2018gain,lu2020multiple}, we also compared the two deep learning models with CART based on the five benchmark datasets and simulation procedure (different from our proposed framework) used in these papers. Details of these datasets and simulations are presented in the supplementary material. The sample sizes of these data are generally not large enough to be considered as population data from which we can repeatedly sample from without replacement, so we are unable to evaluate them in a meaningful way using absolute standardized bias, relative MSE or coverage. We therefore evaluate the methods primarily on the weighted absolute bias metric. In summary, CART again consistently and significantly outperforms MIDA and GAIN in terms of weighted absolute bias for both categorical and continuous variables, across all five benchmark datasets. The difference in performance is particularly pronounced with continuous variables. We also calculated the overall MSE and accuracy as those papers did. Except for one dataset, we could not reproduce the results reported in these papers, even with the authors' code. One possible reason is that the process of tuning and selecting model hyperparameters may not be clearly documented, which is true in the present case. More details are provided in the online supplementary material.

\section{Conclusion} \label{sec:discussion}
Recent years have seen the development of many machine learning based methods for imputing missing data, raising the hope of improving over the more traditional imputation methods such as MICE. However, efforts in evaluating these methods in real world situations remain scarce. In this paper, we adopt an evaluation framework real-data-based simulations. 
We conduct extensive simulation studies based on the American Community Survey to compare repeated sampling properties of two MICE methods and two deep learning imputation methods based on GAN (GAIN) and denoising autoencoders (MIDA). 

We find that the deep learning models hold a vast computational advantage over MICE methods, partially because they can leverage GPU power for high-performance computing. However, our simulations as well as evaluation on several ``benchmark" data suggest that MICE with CART specification of the conditional models consistently outperforms, usually by a substantial margin, the deep learning models in terms of bias, mean squared error, and coverage under a wide range of realistic settings. In particular, GAIN and MIDA tend to generate unstable imputations with enormous variations over repeated samples compared with MICE. One possible explanation is that deep neural networks excel at detecting complex sub-structures of big data, but may not suit for data with simple structure, such as the simulated data used here. Another possibility is that the sample sizes in our simulations are not adequate to train deep neural networks, which usually required much more data compared to traditional statistical models.

These results contradict previous findings based on the single performance metric of overall mean squared error in the machine learning literature \citep[e.g.][]{gondara2017mida, yoon2018gain,lu2020multiple}. This discrepancy highlights the pitfalls of the common practice in the machine learning literature of evaluating imputation methods. It also demonstrates the importance of assessing repeated-sampling properties on multiple estimands of MI methods. An interesting finding is that ensemble trees (e.g. RF)  do not improve over a single tree (e.g. CART) in the context of MICE, which matches the findings in another recent study \citep{wongkamthong2021comparative}. Combined with the fact that the former is more computationally intensive than the latter, we recommend using MICE with CART instead of RF in practice. 

Our study has a few limitations. First, there are many deep learning methods that can be adapted to missing data imputation and all may have different operating characteristics. We choose GAIN and MIDA because both generative adversarial network and denoising autoencoders are immensely popular deep learning methods, and the imputation methods based on them have been advertised as superior to MICE. Nonetheless, it would be desirable to examine other deep learning based imputation methods in future research. Second, performance of machine learning methods is highly dependent on hyperparameter selection. So it can be argued that the inferior performance of GAIN and MIDA may be at least partially due to sub-optimal hyperparameter selection. However, practitioners would most likely rely on default hyperparameter values for any machine learning based imputation methods, which is indeed what we have adopted in our simulations and thus represents the real practice. Third, we did not consider the joint distribution between any categorical and continuous variables, but our evaluations within categorical and continuous variables have yielded consistent conclusions. Lastly, as any simulation study, one should exercise caution in generalizing the conclusions. By carefully selecting the data and metrics, we have attempted to closely mimic the settings representative of real survey data so that our conclusions are informative for practitioners who deal with similar situations. Additional evaluation studies based on different data are desired to shed more insights on the operating characteristics and comparative performances of different missing data imputation methods.  Data, code, and supplementary material for the paper are available at: \url{https://github.com/zhenhua-wang/MissingData_DL}.


\section*{Acknowledgements}
Poulos and Li's research is supported by the National Science Foundation under Grant DMS-1638521 to the Statistical and Applied Mathematical Sciences Institute. 

\renewcommand{\refname}{References}
\printbibliography

@article{de2003prevention,
    title={Prevention and treatment of item nonresponse},
    author={De Leeuw, Edith D and Hox, Joop and Huisman, Mark and others},
    journal={Journal of Official Statistics-Stockholm},
    volume={19},
    number={2},
    pages={153--176},
    year={2003},
    publisher={ALMQVIST \& WIKSELL INTERNATIONAL}
}

@misc{Dua2019,
    author = "Dua, Dheeru and Graff, Casey",
    year = "2017",
  title = {{UCI} {M}achine {L}earning {R}epository},
    url = "http://archive.ics.uci.edu/ml",
    institution = "University of California, Irvine, School of Information and Computer Sciences" }

@inproceedings{glorot2010,
  title={Understanding the Difficulty of Training Deep Feedforward Neural Networks.},
  author={Glorot, Xavier and Bengio, Yoshua},
  booktitle={Artificial Intelligence and Statistics},
  volume={9},
  pages={249--256},
  year={2010}
}

@article{kingma2014adam,
  title={Adam: A Method for Stochastic Optimization},
  author={Kingma, Diederik and Ba, Jimmy},
  journal={arXiv:1412.6980},
  year={2014}
}

@book{little2014,
  title={Statistical Analysis with Missing Data},
  author={Little, Roderick JA and Rubin, Donald B},
  year={2014},
  publisher={Hoboken, NJ: John Wiley \& Sons}
}

@article{yoon2018estimating,
  title={Estimating missing data in temporal data streams using multi-directional recurrent neural networks},
  author={Yoon, Jinsung and Zame, William R and van der Schaar, Mihaela},
  journal={IEEE Transactions on Biomedical Engineering},
  volume={66},
  number={5},
  pages={1477--1490},
  year={2018},
  publisher={IEEE}
}

@inproceedings{yoon2018gain,
  title={{Gain}: Missing data imputation using generative adversarial nets},
  author={Yoon, Jinsung and Jordon, James and Schaar, Mihaela},
  booktitle={International conference on machine learning},
  pages={5689--5698},
  year={2018},
  organization={PMLR}
}

@inproceedings{gondara2017mida,
  title={{MIDA}: Multiple imputation using denoising autoencoders},
  author={Gondara, Lovedeep and Wang, Ke},
  booktitle={Pacific-Asia Conference on Knowledge Discovery and Data Mining},
  pages={260--272},
  year={2018},
  organization={Springer}
}

@inproceedings{monti2017geometric,
  title={Geometric matrix completion with recurrent multi-graph neural networks},
  author={Monti, Federico and Bronstein, Michael and Bresson, Xavier},
  booktitle={Advances in Neural Information Processing Systems},
  pages={3697--3707},
  year={2017}
}

@inproceedings{fortuin2019gpvae,
  title={{GP-VAE}: Deep probabilistic time series imputation},
  author={Fortuin, Vincent and Baranchuk, Dmitry and R{\"a}tsch, Gunnar and Mandt, Stephan},
  booktitle={International Conference on Artificial Intelligence and Statistics},
  pages={1651--1661},
  year={2020},
  organization={PMLR}
}

@article{che2018recurrent,
  title={Recurrent neural networks for multivariate time series with missing values},
  author={Che, Zhengping and Purushotham, Sanjay and Cho, Kyunghyun and Sontag, David and Liu, Yan},
  journal={Scientific Reports},
  volume={8},
  number={1},
  pages={1--12},
  year={2018},
  publisher={Nature Publishing Group}
}

@inproceedings{cao2018brits,
  title={{BRITS}: Bidirectional recurrent imputation for time series},
  author={Cao, Wei and Wang, Dong and Li, Jian and Zhou, Hao and Li, Lei and Li, Yitan},
  booktitle={Advances in Neural Information Processing Systems},
  pages={6775--6785},
  year={2018}
}

@article{lipton2016modeling,
  title={Modeling missing data in clinical time series with {RNN}s},
  author={Lipton, Zachary C and Kale, David C and Wetzel, Randall},
  journal={Machine Learning for Healthcare},
  volume={56},
  year={2016}
}

@article{akande2017empirical,
  title={An empirical comparison of multiple imputation methods for categorical data},
  author={Akande, Olanrewaju and Li, Fan and Reiter, Jerome},
  journal={The American Statistician},
  volume={71},
  number={2},
  pages={162--170},
  year={2017},
  publisher={Taylor \& Francis}
}

@inproceedings{goodfellow2014gan,
  title={Generative adversarial nets},
  author={Goodfellow, Ian and Pouget-Abadie, Jean and Mirza, Mehdi and Xu, Bing and Warde-Farley, David and Ozair, Sherjil and Courville, Aaron and Bengio, Yoshua},
  booktitle={Advances in Neural Information Processing Systems},
  pages={2672--2680},
  year={2014}
}

@book{LittleRubin2019,
  title={Statistical Analysis with Missing Data, 3rd Edition},
  author={Little, Roderick JA and Rubin, Donald B},
  year={2019},
  publisher = {New York: John Wiley \& Sons}
}

@article{Rubin1976,
    Journal = {Biometrika},
    Volume = 63,
    Pages = {581--592},
    author = {Rubin, D. B.},
    title = {Inference and Missing data (with discussion)},
    Year = 1976
}

@book{Rubin1987,
author = {Rubin, D. B.},
title = {Multiple Imputation for Nonresponse in Surveys},
Year = 1987,
Publisher = {New York: John Wiley \& Sons},
Pages = {258}
}

@article{VanBuuren2006,
author = {van {B}uuren, S. and Brand, J. P. L. and Groothuis-Oudshoorn, C. G. M. and Rubin, D. B.},
title = {Fully conditional specification in multivariate imputation},
journal = {Journal of Statistical Computation and Simulation},
volume = {76},
number = {12},
pages = {1049-1064},
year  = {2006},
publisher = {Taylor & Francis}
}

@article{GelmanSpeed1993,
  author = {Gelman, A. and Speed, T. P},
  title = {Characterizing a joint probability distribution by conditionals},
  journal = {Journal of the Royal Statistical Society Series B: Statistical Methodology},
  year = {1993},
  volume = {55},
  pages = {185--188}
}

@article{ArnoldPress1989,
  author = {Arnold, B. C. and Press, S. J.},
  title = {Compatible Conditional Distributions},
  journal = {Journal of the American Statistical Association},
  year = {1989},
  volume = {84},
  pages = {152--156}
}

@TECHREPORT{Li2012,
  author = {Li, F and Yu, Y and Rubin, DB},
  title = {Imputing missing data by fully conditional models: Some cautionary examples and guidelines},
  institution = {Duke University Department of Statistical Science Discussion Paper 11-24},
  year = {2012},
  pages = {1-35}
}

@article{lu2020multiple,
  title={Multiple imputation with denoising autoencoder using metamorphic truth and imputation feedback},
  author={Lu, Haw-minn and Perrone, Giancarlo and Unpingco, Jos{\'e}},
  journal={arXiv preprint arXiv:2002.08338},
  year={2020}
}

@book{breiman:1984,
author = {Breiman, L. and Friedman, J. H. and Olshen, R. A. and Stone, C. J.},
title = {Classification and Regression Trees},
Year = 1984,
Publisher = {Belmont, CA: Wadsworh, Inc.}
}

@article{burgreit10,
author = {Burgette, L. and Reiter, J. P.},
Journal = {American Journal of Epidemiology},
Year = {2010},
title = {Multiple imputation via sequential regression trees},
Pages = {1070--1076},
Volume = 172}

@article{breiman:2001,
author = {Breiman, L.},
Journal = {Machine Learning},
title = {Random forests},
Year = 2001,
Volume= 45,
Pages = {5--32}}

@article{Shah:2014,
author = {Shah, Anoop and Bartlett, Jonathan and Carpenter, James and Nicholas, Owen and Hemingway, Harry},
year = {2014},
month = {03},
pages = {764-74},
title = {Comparison of random forest and parametric imputation models for imputing missing data using MICE: A CALIBER study},
volume = {179},
journal = {American Journal of Epidemiology},
%doi = {10.1093/aje/kwt312}
}

@article {white2011,
author = {White, Ian R. and Royston, Patrick and Wood, Angela M.},
title = {Multiple imputation using chained equations: Issues and guidance for practice},
journal = {Statistics in Medicine},
volume = {30},
number = {4},
publisher = {John Wiley & Sons, Ltd.},
issn = {1097-0258},
%url = {http://dx.doi.org/10.1002/sim.4067},
%doi = {10.1002/sim.4067},
pages = {377--399},
keywords = {missing data, multiple imputation, fully conditional specification},
year = {2011},
}

@book{van2018flexible,
  title = {Flexible Imputation of Missing Data},
  author = {van Buuren, S.},
  isbn = {9781138588318},
  lccn = {2018017122},
  series = {Chapman \& Hall/CRC Interdisciplinary Statistics},
  year = {2018},
  publisher = {CRC Press LLC}
}

@article{Rubin1996,
  title={Multiple imputation after 18+ years},
  author={Rubin, Donald B},
  journal={Journal of the American Statistical Association},
  volume={91},
  number={434},
  pages={473--489},
  year={1996},
  publisher={Taylor \& Francis Group}
}

@article{BarnardMeng1999,
  title={Applications of multiple imputation in medical studies: from {AIDS} to {NHANES}},
  author={Barnard, John and Meng, Xiao-Li},
  journal={Statistical Methods in Medical Research},
  volume={8},
  number={1},
  pages={17--36},
  year={1999},
  publisher={Sage Publications Sage CA: Thousand Oaks, CA}
}

@article{ReiterRaghunathan2007,
  title={The multiple adaptations of multiple imputation},
  author={Reiter, Jerome P and Raghunathan, Trivellore E},
  journal={Journal of the American Statistical Association},
  volume={102},
  number={480},
  pages={1462--1471},
  year={2007},
  publisher={Taylor \& Francis}
}

@article{HarelZhou2007,
  title={Multiple imputation: {R}eview of theory, implementation and software},
  author={Harel, Ofer and Zhou, Xiao-Hua},
  journal={Statistics in Medicine},
  volume={26},
  number={16},
  pages={3057--3077},
  year={2007},
  publisher={Wiley Online Library}
}

@book{Schafer1997a,
  title={Analysis of Incomplete Multivariate Data},
  author={Schafer, Joseph L},
  year={1997},
  publisher={Chapman \& Hall: London}
}

@inproceedings{Vincent2008,
  title={Extracting and composing robust features with denoising autoencoders},
  author={Vincent, Pascal and Larochelle, Hugo and Bengio, Yoshua and Manzagol, Pierre-Antoine},
  booktitle={Proceedings of the 25th international conference on Machine learning},
  pages={1096--1103},
  year={2008}
}

@article{missforest,
  title={MissForest—non-parametric missing value imputation for mixed-type data},
  author={Stekhoven, Daniel J and B{\"u}hlmann, Peter},
  journal={Bioinformatics},
  volume={28},
  number={1},
  pages={112--118},
  year={2012},
  publisher={Oxford University Press}
}

@INPROCEEDINGS{randomForest,
  author={ {Tin Kam Ho}},
  booktitle={Proceedings of 3rd International Conference on Document Analysis and Recognition}, 
  title={Random decision forests}, 
  year={1995},
  volume={1},
  number={},
  pages={278-282 vol.1}
  }

@inproceedings{maas2013rectifier,
  title={Rectifier nonlinearities improve neural network acoustic models},
  author={Maas, Andrew L and Hannun, Awni Y and Ng, Andrew Y},
  booktitle={Proc. {ICML}},
  series={30},
  number={1},
  pages={3},
  year={2013}
}

@article{LiEtAl2014:JCGS,
author = {Fan Li and Michela Baccini and Fabrizia Mealli and Elizabeth R. Zell and Constantine E. Frangakis and Donald B. Rubin},
title = {Multiple Imputation by Ordered Monotone Blocks With Application to the Anthrax Vaccine Research Program},
journal = {Journal of Computational and Graphical Statistics},
volume = {23},
number = {3},
pages = {877-892},
year  = {2014},
publisher = {Taylor & Francis}
}

@misc{tensorflow2015-whitepaper,
title={ {TensorFlow}: Large-Scale Machine Learning on Heterogeneous Systems},
url={https://www.tensorflow.org/},
note={Software available from tensorflow.org},
author={
    Martín~Abadi and
    Ashish~Agarwal and
    Paul~Barham and
    Eugene~Brevdo and
    Zhifeng~Chen and
    Craig~Citro and
    Greg~S.~Corrado and
    Andy~Davis and
    Jeffrey~Dean and
    Matthieu~Devin and
    Sanjay~Ghemawat and
    Ian~Goodfellow and
    Andrew~Harp and
    Geoffrey~Irving and
    Michael~Isard and
    Yangqing Jia and
    Rafal~Jozefowicz and
    Lukasz~Kaiser and
    Manjunath~Kudlur and
    Josh~Levenberg and
    Dandelion~Mané and
    Rajat~Monga and
    Sherry~Moore and
    Derek~Murray and
    Chris~Olah and
    Mike~Schuster and
    Jonathon~Shlens and
    Benoit~Steiner and
    Ilya~Sutskever and
    Kunal~Talwar and
    Paul~Tucker and
    Vincent~Vanhoucke and
    Vijay~Vasudevan and
    Fernanda~Viégas and
    Oriol~Vinyals and
    Pete~Warden and
    Martin~Wattenberg and
    Martin~Wicke and
    Yuan~Yu and
    Xiaoqiang~Zheng},
  year={2015},
}

@article{tang2005comparison,
  title={A comparison of imputation methods in a longitudinal randomized clinical trial},
  author={Tang, Lingqi and Song, Juwon and Belin, Thomas R and Un{\"u}tzer, J{\"u}rgen},
  journal={Statistics in medicine},
  volume={24},
  number={14},
  pages={2111--2128},
  year={2005},
  publisher={Wiley Online Library}
}

@article{huque2018comparison,
  title={A comparison of multiple imputation methods for missing data in longitudinal studies},
  author={Huque, Md Hamidul and Carlin, John B and Simpson, Julie A and Lee, Katherine J},
  journal={BMC medical research methodology},
  volume={18},
  number={1},
  pages={1--16},
  year={2018},
  publisher={BioMed Central}
}

@Article{mice2011,
    title = {{mice}: Multivariate Imputation by Chained Equations in
      R},
    author = {van Buuren, Stef and Karin Groothuis-Oudshoorn},
    journal = {Journal of Statistical Software},
    year = {2011},
    volume = {45},
    number = {3},
    pages = {1-67},
    %url = {https://www.jstatsoft.org/v45/i03/},
  }

@article{manrique2014bayesian,
  title={Bayesian estimation of discrete multivariate truncated latent structure models},
  author={Manrique-Vallier, D and Reiter, J},
  journal={Journal of Computational and Graphical Statistics},
  year={2014},
  volume={23}, 
  pages={1061–1079}
}

@article{murray2016multiple,
  title={Multiple imputation of missing categorical and continuous values via Bayesian mixture models with local dependence},
  author={Murray, Jared S and Reiter, Jerome P},
  journal={Journal of the American Statistical Association},
  volume={111},
  number={516},
  pages={1466--1479},
  year={2016},
  publisher={Taylor \& Francis}
}

@article{raghunathan2001multivariate,
  title={A multivariate technique for multiply imputing missing values using a sequence of regression models},
  author={Raghunathan, Trivellore E and Lepkowski, James M and Van Hoewyk, John and Solenberger, Peter},
  journal={Survey Methodology},
  volume={27},
  number={1},
  pages={85--96},
  year={2001}
}

@article{royston2011multiple,
  title={Multiple imputation by chained equations (MICE): implementation in {S}tata},
  author={Royston, Patrick and White, Ian R},
  journal={J Statistical Software},
  volume={45},
  number={4},
  pages={1--20},
  year={2011}
}

@article{honaker2011amelia,
  title={{Amelia II}: A program for missing data},
  author={Honaker, James and King, Gary and Blackwell, Matthew},
  journal={Journal of Statistical Software},
  volume={45},
  number={7},
  pages={1--47},
  year={2011}
}

@article{yuan2011multiple,
  title={Multiple imputation using {SAS} software},
  author={Yuan, Yang},
  journal={Journal of Statistical Software},
  volume={45},
  number={6},
  pages={1--25},
  year={2011}
}

@article{su2011multiple,
  title={Multiple imputation with diagnostics (mi) in R: Opening windows into the black box},
  author={Su, Yu-Sung and Gelman, Andrew E and Hill, Jennifer and Yajima, Masanao},
  year={2011},
  journal={Journal of Statistical Software},
  volume={45},
}

@article{DooverEtAl2014,
title = {Recursive partitioning for missing data imputation in the presence of interaction effects},
journal = {Computational Statistics \& Data Analysis},
volume = {72},
pages = {92-104},
year = {2014},
issn = {0167-9473},
%doi = {https://doi.org/10.1016/j.csda.2013.10.025},
author = {L.L. Doove and S. {Van Buuren} and E. Dusseldorp}
}

@book{HastieESL:2009,
  author = {Hastie, Trevor and Tibshirani, Robert and Friedman, Jerome},
  edition = 2,
  publisher = {Springer},
  title = {The elements of statistical learning: data mining, inference and prediction},
  year = 2009
}

@article{vincent2010stacked,
  title={Stacked denoising autoencoders: Learning useful representations in a deep network with a local denoising criterion.},
  author={Vincent, Pascal and Larochelle, Hugo and Lajoie, Isabelle and Bengio, Yoshua and Manzagol, Pierre-Antoine and Bottou, L{\'e}on},
  journal={Journal of Machine Learning Research},
  volume={11},
  number={12},
  year={2010}
}

@article{horton2003potential,
  title={A potential for bias when rounding in multiple imputation},
  author={Horton, Nicholas J and Lipsitz, Stuart R and Parzen, Michael},
  journal={The American Statistician},
  volume={57},
  number={4},
  pages={229--232},
  year={2003},
  publisher={Taylor \& Francis}
}

@article{chen2019recent,
  title={Recent developments in dealing with item non-response in surveys: a critical review},
  author={Chen, Sixia and Haziza, David},
  journal={International Statistical Review},
  volume={87},
  pages={S192--S218},
  year={2019},
  publisher={Wiley Online Library}
}

@article{haziza2020variance,
  title={Variance estimation procedures in the presence of singly imputed survey data: a critical review},
  author={Haziza, David and Vall{\'e}e, Audrey-Anne},
  journal={Japanese Journal of Statistics and Data Science},
  volume={3},
  number={2},
  pages={583--623},
  year={2020},
  publisher={Springer}
}

@article{wongkamthong2021comparative,
  title={A Comparative Study of Imputation Methods for Multivariate Ordinal Data},
  author={Wongkamthong, Chayut and Akande, Olanrewaju},
  journal={Journal of Survey Statistics and Methodology},
  volume={in press},
  %arXiv preprint arXiv:2010.10471
  year={2021}
}

@article{DBLP:journals/corr/BerthelotSM17,
  author    = {David Berthelot and
               Tom Schumm and
               Luke Metz},
  title     = {{BEGAN:} Boundary Equilibrium Generative Adversarial Networks},
  journal   = {CoRR},
  volume    = {abs/1703.10717},
  year      = {2017},
  url       = {http://arxiv.org/abs/1703.10717},
  eprinttype = {arXiv},
  eprint    = {1703.10717},
  bibsource = {dblp computer science bibliography, https://dblp.org}
}

@article{ham2020unbalanced,
  title={Unbalanced gans: Pre-training the generator of generative adversarial network using variational autoencoder},
  author={Ham, Hyungrok and Jun, Tae Joon and Kim, Daeyoung},
  journal={arXiv preprint arXiv:2002.02112},
  year={2020}
}

\end{document}


\maketitle
\thispagestyle{empty}

\begin{abstract}
  This supplementary material provides pseudocode for the MICE algorithm, and includes the data dictionary for the variables used in the simulation studies in the main text. It also includes the nonresponse models used to create the MAR scenario on the main text. Finally, it includes an evaluation based on five well-studied ``benchmark'' datasets.
\end{abstract}

\newpage
\setcounter{page}{1}

\section{MICE algorithm}

	\begin{algorithm}[H]
		\begin{algorithmic}[1]
			\item[ ] {\textbf{input:}} incomplete data  ${\textbf Y} = ({\textbf Y}_{\textrm{obs}}, {\textbf Y}_{\textrm{mis}})$
			\item[ ] {\textbf{output:}} completed data ${\textbf Y}^{(l)} = ({\textbf Y}_{\textrm{obs}}, {\textbf Y}_{\textrm{mis}}^{(T)})$
			\STATE Specify order $(\textbf{Y}_{(1)}, \dots, \textbf{Y}_{(p)})$ to iterate through sequence of conditional models
			\FOR{$l=1$ to $L$}
			\STATE initialize each ${\textbf Y}_{\textrm{mis},(j)}$ by sampling from ${\textbf Y}_{\textrm{obs},(j)}$
			\FOR{$t=1$ to $T$}
			\FOR{$j=1$ to $p$}
			\STATE fit conditional model\\ $(\textbf{Y}_{(j)} | \textbf{Y}_{\textrm{obs},(j)}, \{\textbf{Y}_{(k)}^{(t)}: k<j\}, \{\textbf{Y}_{(k)}^{(t-1)}: k>j\})$
			\STATE replace ${\textbf Y}_{\textrm{mis},(j)}^{(t)}$ with draws from implied model\\ $({\textbf Y}_{\textrm{mis},(j)}^{(t)} | \textbf{Y}_{\textrm{obs},(j)}, \{\textbf{Y}_{(k)}^{(t)}: k<j\}, \{\textbf{Y}_{(k)}^{(t-1)}: k>j\})$
			\ENDFOR
			\ENDFOR
			\ENDFOR
		\end{algorithmic}
		\caption{MICE}
		\label{alg:mice}
	\end{algorithm}
	
\section{Data dictionary for the ACS application}


First, we present the data dictionary for the variables used in the simulation scenarios in Section 4 of the main paper. After pre-processing the 2018 American Community Survey (ACS) data as discussed in the main paper, we have 1,257,501 units, with 18 binary variables, 20 categorical variables with 3 to 9 levels, and 8 continuous variables. We treat this processed version as our final population from which we repeatedly sample from. Table \ref{table:vars} describes variables in this final population data.
\begin{table}
    \caption{Variables from the 2018 ACS used to construct our population data.} 
    \centering
    \footnotesize
    \begin{tabular}{lll}
        \toprule
        Variable & Description & Type \\
        \midrule
        ACR & Lot size & Ordinal (3 levels)\\
        AGEP & Age & Numeric\\
        BDSP & Number of bedrooms & Numeric\\
        BLD & Units in structure & Nominal (4 levels)\\
        COW & Class of worker & Nominal (4 levels)\\
        DIS & Disability & Binary\\
        HHL & Household language & Nominal (3 levels)\\
        HHT & HH/family type & Nominal (4 levels)\\
        HINS1 & Insurance through a current or former employer or union & Binary\\
        HINS2 & Insurance purchased directly from an insurance company & Binary\\
        HINS3 & Medicare for people $>=65$ years or with certain disabilities & Binary \\
        HUGCL & Household with grandparent living with grandchildren & Binary\\
        HUPAC & HH presence and age of children & Nominal (4 levels)\\
        HUPAOC & HH presence and age of own children & Nominal (4 levels)\\
        HUPARC & HH presence and age of related children & Nominal (4 levels)\\
        JWMNP & Travel time to work & Numeric\\
        JWRIP & Vehicle occupancy & Nominal (3 levels)\\
        JWTR & Means of transportation to work & Nominal (3 levels)\\
        LAPTOP & Laptop or desktop & Binary\\
        LNGI & Limited English speaking household & Binary\\
        MAR & Marital status & Nominal (4 levels)\\
        MARHT & Number of times married & Ordinal (4 levels)\\
        MULTG & Multigenerational household & Binary\\
        MV & When moved into this house or apartment & Ordinal (7 levels)\\
        NP & Number of persons in this household & Numeric\\
        NR & Presence of nonrelative in household & Binary\\
        PAOC & Presence and age of own children & Nominal (3 levels)\\
        PARTNER & Unmarried partner household & Binary\\
        PRIVCOV & Private health insurance coverage recode & Binary\\
        PSF & Presence of subfamilies in household & Binary\\
        PUBCOV & Public health coverage recode & Binary \\
        PWGTP & Person's weight & Numeric \\
        R18 & Presence of persons under 18 years in household & Binary\\
        R65 & Presence of persons 65 years and over in household & Ordinal (3 levels)\\
        RAC1P & Recoded detailed race code & Binary\\
        RMSP & Number of rooms & Numeric\\
        SCHL & Educational attainment & Ordinal (8 levels)\\
        SEX & Sex & Binary\\
        SRNT & Specified rental unit & Binary\\
        SVAL & Specified owner unit & Binary\\
        TEN & Tenure & Nominal (3 levels)\\
        VEH & Vehicles (1 ton or less) available & Ordinal (4 levels)\\
        WAGP & Wages or salary income past 12 months  & Numeric\\
        WIF & Workers in family during the past 12 months & Ordinal (4 levels)\\
        WKHP & Usual hours worked per week past 12 months & Numeric\\
        YBL & When structure first built & Ordinal (9 levels)\\
        \bottomrule
    \end{tabular}
    \label{table:vars} 
\end{table}




\section{Nonresponse models for MAR scenario}

Next, we describe the MAR scenario in the main paper in more detail. We begin by setting six variables --- age, sex, marital status, race, educational attainment and class of worker --- to be fully observed. 
We then randomly split the remaining 40 variables into three groups, consisting of 10, 15, and 15 variables. We create one logistic model per group, and condition on two of the fully observed six variables within each model. We use the logistic models to generate the item nonresponse indicators for each variable in that group.

Specifically, let $A_i$, $S_i$, $M_i$, $R_i$, $E_i$ and $C_i$  represent the age, sex, marital status, race, educational attainment and class of worker, of the $i=1, \dots, n$ individuals in the data. For each variable in the first group of 10 variables, we sample the item nonresponse indicator for each individual $i$ using probability $p_{i1}$. For the second and third groups of 15 variables each, we use $p_{i2}$ and $p_{i3}$ respectively. We define $p_{i1}$, $p_{i2}$, and $p_{i2}$, as follows.
\begin{align}
&\begin{aligned}
\text{logit}(p_{i1}) = & -1.3 + 0.4 \cdot \mathds{1}[C_i=0] - 1.3 \cdot \mathds{1}[C_i=1]  \\
& + 1.2 \cdot \mathds{1}[C_i=2] -0.4 \cdot \mathds{1}[C_i=3] + 0.02 \cdot A_i;\\
\end{aligned}\\
&\begin{aligned}
\text{logit}(p_{i2}) = & -1.2 -1.5 \cdot \mathds{1}[E_i \in \{0,1\}] + 0.5 \cdot \mathds{1}[E_i \in \{2,3,4\}] \\
& + \mathds{1}[E_i \in \{5,6,7\}] - 0.3 \cdot \mathds{1}[M_i=0] + 1.2 \cdot \mathds{1}[M_i=1] \\
& - 1.3 \cdot \mathds{1}[M_i=2] + 0.4 \cdot \mathds{1}[M_i=3];\\
\end{aligned}\\
&\begin{aligned}
\text{logit}(p_{i3}) = & -0.5 - 0.7 \cdot \mathds{1}[S_i=0] + 0.9 \cdot \mathds{1}[S_i=1]  \\
& - 1.5 \cdot \mathds{1}[R_i=0] + 0.6 \cdot \mathds{1}[R_i=1].
\end{aligned}
\end{align}
Here, $\mathds{1}[\cdot]=1$  when its argument is true and $\mathds{1}[\cdot]=0$ otherwise. Across all three groups, we set the corresponding entries for each variables as missing when the sampled nonresponse indicator is equal to one. This process results in approximately 30\% missing rate for each of the 40 variables.

\section{Evaluation based on ``benchmark'' datasets} \label{sec:ml-eval}
Finally, we check the reproducibility and verify the claims of the deep learning methods, by comparing the methods based on our metrics using five ``benchmark'' datasets commonly used in the machine learning literature. The benchmark datasets are from the UCI Machine Learning Repository \citep{Dua2019}, and are frequently used in the machine learning literature for evaluating missing data imputation methods. These datasets, which we describe in Table \ref{tab:benchmark-characteristics}, vary vastly in size and structure. For example, the Breast Cancer dataset has only 569 sample units and no categorical variables; the Spam Detection dataset contains 4,601 observations but only continuous variables, whereas the Letter Recognition dataset contains 20,000 observations but only categorical variables. These extremes are very different from what one would expect from a survey data, for example, the ACS data used in our main evaluations.

For each dataset, we follow \citet{yoon2018estimating} and create an MCAR scenario by randomly setting 20\% of the values of each variable to be missing independently. We use MICE-CART, GAIN, and MIDA to create $L = 10$ completed datasets for each benchmark dataset, and compute the metrics on the completed datasets. We again omit MICE-RF because the results in the main text already showed MICE-RF to be consistently inferior to MICE-CART in terms of performance and computation. We only generate 10 samples to remain consistent with the other scenarios in the main paper.
\begin{table}
\caption{\label{tab:benchmark-characteristics}Characteristics of selected UCI benchmark datasets.}
\centering
\footnotesize
\begin{threeparttable}
\begin{tabular}{lrrrr}
\toprule
Data & $\#$ Samples & $\#$ Cont. variables & $\#$ Cat. variables & Avg. Corr$^{\star}$ \\
\midrule
Breast Cancer & 569 & 30 & 0 & 0.394 \\
Credit Card Default & 30,000 & 14 & 9 & 0.163 \\
Letter Recognition & 20,000 & 0 & 16 & 0.182 \\
Spam Detection & 4,601 & 57 & 0 & 0.060 \\
News Shares & 39,644 & 44 & 14 & 0.068 \\
\bottomrule
\end{tabular}
\begin{tablenotes}
\item[] $^\star$ is the average absolute correlations among variables, as reported in \citet{yoon2018estimating}.
\end{tablenotes}
\end{threeparttable}
\end{table}

Also, for consistency, we prefer to evaluate all five datasets in the same manner. However, the sample sizes are not large enough to consider them population data from which we can repeatedly sample from without replacement. For example, the Breast Cancer dataset only contains 569 observations and the Spam Detection dataset only contains 4,601. Thus, we are unable to evaluate them in a meaningful way using absolute normalized bias, relative MSE or coverage. We therefore primarily evaluate them in a similar manner to the n=100,000 and 30\% MCAR scenario in the main text. That is, we evaluate them using the weighted absolute bias metric. 

We do report estimates of overall RMSE on the continuous variables and overall accuracy on the categorical variables, to once again provide evidence on how examining only the overall RMSE and accuracy metrics may be misleading for evaluating imputation methods. Here, we repeatedly create the same 20\% MCAR process on each dataset, 10 times, and estimate the overall RMSE and accuracy on all 10 copies of the data. We report the average overall RMSE and accuracy across the 10 copies, and report the standard deviation of the values as the corresponding standard errors. While we once again prefer to draw repeated samples from a population as a way to properly account for the sampling mechanism, we follow this approach of repeatedly creating missing values of a single dataset to replicate the results from \citet{yoon2018estimating} for these two metrics on all the datasets, as much as possible.

Table \ref{table:weighted_absdiff_uci} displays the median values of the estimated weighted absolute bias of the marginal probabilities of the categorical and binned continuous variables. MICE-CART significantly outperforms the other two methods. Specifically, MICE-CART results in the smallest weighted absolute bias in both categorical and continuous variables, across all the datasets. The difference is more pronounced with continuous variables, particularly with the Spam Detection dataset, which has 57 continuous variables, the highest number of all the datasets. This result is consistent with our findings in the main paper.
\begin{table}
\caption{\label{table:weighted_absdiff_uci}Median values of the weighted absolute bias $(\times 100)$, of the marginal probabilities of the categorical and binned continuous variables.}
\centering
\footnotesize
\begin{tabular}{lrrr}
\toprule
Data & MICE-CART & GAIN & MIDA \\
 \midrule
\vspace{-10pt} \\
\multicolumn{4}{c}{\underline{Categorical}}\\
Credit Card Default     & 0.07 & 1.76 & 0.56\\
News Shares             & 0.09 & 4.24 & 1.14\\
Letter Recognition      & 0.08 & 2.31 & 0.39\\
\midrule
\vspace{-10pt} \\
\multicolumn{4}{c}{\underline{Binned Continuous}}\\
Breast Cancer          & 0.50 & 1.30  & 0.96\\
Credit Card Default    & 0.06 & 3.19  & 1.79\\
News Shares            & 0.07 & 3.69  & 3.16\\
Spam Detection         & 0.18 & 14.86 & 15.08\\
\bottomrule
\end{tabular}
\end{table}

Table \ref{table:accuracy _uci} displays the overall RMSE on continuous variables and overall accuracy on categorical variables for the datasets. MICE-CART achieves the highest overall accuracy in the Credit Card Default and News Shares datasets, while MIDA achieves the highest accuracy in the Letter Recognition dataset. There is no consistent pattern in comparing all three methods using overall RMSE. MICE-CART achieves the lowest overall RMSE in the News Shares dataset, MIDA achieves the lowest overall RMSE in the Credit Card Default dataset, GAIN achieves the lowest overall RMSE in the Breast Cancer dataset, and both GAIN and MIDA achieve similar RMSE, lower than MICE-CART, in the Spam Detection dataset. MIDA is capable of preserving feature correlations \citep{gondara2017mida} and performs comparatively well against the other methods on the datasets with highly correlated features. These patterns clearly differ from those reported earlier based on marginal and bivariate probabilities and weighted absolute bias. Therefore, as discussed in the main paper, we again warn against using the overall RMSE and accuracy as the only metrics for comparing imputation methods.
\begin{table}
\caption{\label{table:accuracy _uci}Overall RMSE on continuous variables and overall accuracy on categorical variables for each dataset, with estimated standard errors.}
\centering
\footnotesize
\begin{tabular}{lrrr}
\toprule
Data & MICE-CART & GAIN & MIDA \\
\midrule
\vspace{-10pt} \\
\multicolumn{4}{c}{\underline{RMSE}}\\
Breast Cancer          & 0.107 $\pm$ 0.002 & 0.078 $\pm$ 0.003  & 0.100 $\pm$ 0.003\\
Credit Card Default    & 0.073 $\pm$ 0.003 & 0.075 $\pm$ 0.006  & 0.067 $\pm$ 0.003\\
News Shares            & 0.121 $\pm$ 0.021 & 0.181 $\pm$ 0.014  & 0.173 $\pm$ 0.015\\
Spam Detection         & 0.064 $\pm$ 0.001 & 0.054 $\pm$ 0.001  & 0.055 $\pm$ 0.001\\
 \midrule
\vspace{-10pt} \\
\multicolumn{4}{c}{\underline{Accuracy}}\\
Credit Card Default     & 0.740 $\pm$ 0.001 & 0.642 $\pm$ 0.031 & 0.678 $\pm$ 0.002\\
News Shares             & 0.930 $\pm$ 0.001 & 0.796 $\pm$ 0.053 & 0.876 $\pm$ 0.001\\
Letter Recognition      & 0.395 $\pm$ 0.001 & 0.332 $\pm$ 0.015 & 0.443 $\pm$ 0.001\\
\bottomrule
\end{tabular}
\end{table}

Finally, we note that we are unable to perfectly replicate the results from \citet{yoon2018estimating} on all the datasets. First, for datasets containing both categorical and continuous variables, that is the Credit Card Default and News Shares datasets, the authors do not report a separate overall RMSE for the continuous variables as we do here. The authors also primarily report overall RMSE for the Letter Recognition dataset, while we report overall accuracy since the dataset only contains categorical variables. We do replicate the results for the Spam Detection dataset, and while we are unable to perfectly replicate the results for the Breast Cancer dataset, the overall trends in \citet{yoon2018estimating} are consistent with our results here. We note that the Spam Detection and Breast Cancer datasets contain only continuous variables. Second, the publicly available version of the author's code for GAIN is more updated than what was used in the \citet{yoon2018estimating}. We also did not find any publicly available version of the author's code for MIDA used in \citet{lu2020multiple}; we simply reproduced the architecture. Thus, another possible reason for the discrepancies is the differences in parameter tuning. As we mentioned in the main paper, this once again highlights how much the performance of machine learning methods can be highly dependent on parameter tuning.

\renewcommand{\refname}{References}
\printbibliography